\documentclass{article}
\usepackage{iclr2025_conference,times}

\usepackage{amsmath}
\usepackage{amssymb}
\usepackage{natbib}
\usepackage{float}
\usepackage{tabu}
\usepackage{algorithm}
\usepackage{tikz}
\RequirePackage{algorithmic}
\usepackage{fontawesome5}
\usepackage{hyperref}
\usepackage{url}
\usepackage[utf8]{inputenc} 
\usepackage{url}           
\usepackage{booktabs}       
\usepackage{amsfonts}       
\usepackage{nicefrac}       
\usepackage{microtype}      
\usepackage{xcolor}         
\usepackage{tcolorbox}

\usepackage{colortbl}

\usepackage[utf8]{inputenc} 
\usepackage[T1]{fontenc}    
\usepackage{hyperref}
\usepackage{url}            
\usepackage{booktabs}       
\usepackage{amsfonts}       
\usepackage{nicefrac}       
\usepackage{microtype}      
\usepackage{xcolor}
\usepackage{tabularx}
\usepackage{amsmath}
\usepackage{multirow}
\usepackage{array}
\usepackage{caption}
\usepackage{wrapfig}
\usepackage{enumitem}
\usepackage{lipsum}
\usepackage{subcaption}
\usepackage{multirow} 
\usepackage{adjustbox}

\usepackage{amssymb}
\usepackage{unicode}  
\usepackage[capitalize]{cleveref} 

\usepackage{pgf}
\usepackage{colortbl}

\usepackage{tipa}

\usepackage{tcolorbox}
\usepackage{rotating}
\usepackage[abs]{overpic}
\usepackage{makecell}
\usepackage{longtable}

\usepackage{tocloft}  

\usepackage[english]{babel}
\usepackage{csquotes}
\usepackage{graphicx}
\usepackage{longtable}
\usepackage{caption}
\usepackage{mdframed}
\usepackage{subcaption}
\usepackage{multirow}
\usepackage{placeins}
\usepackage{multicol}
\usepackage{makecell}
\usepackage{soul}
\usepackage[normalem]{ulem} 
\usepackage{wrapfig}
\usepackage{lipsum}
\usepackage{wrapfig}
\usepackage{csquotes}
\usepackage[OT2,T1]{fontenc}
\usepackage[english]{babel}
\usepackage{devanagari}
\usepackage{tablefootnote}
\usepackage{pdfpages}
\usepackage{colortbl}
\usepackage{arydshln}
\usepackage{array}
\usepackage{enumitem}
\usepackage{epigraph} 

\definecolor{myblue}{RGB}{0, 0, 255}
\definecolor{mygray}{gray}{0.9}
\hypersetup{colorlinks=true,breaklinks,linkcolor=red,citecolor=goodblue}

\newcommand{\eg}{\textit{e}.\textit{g}.}

\usepackage{pifont}

\newcommand{\cmark}{\ding{52}}
\newcommand{\fmark}{\ding{56}}
\def\halfcheckmark{\textcolor{black}{\ding{52}}{\small\textcolor{black}{\kern-0.7em\ding{55}}}}

\newcommand{\authorskip}{\hspace{4.8mm}}

\newcommand{\myparagraph}[1]{{\bf #1}}

\definecolor{myGreen}{rgb}{0, .6, .0}
\definecolor{goodblue}{HTML}{0071bc}

\usepackage[capitalize]{cleveref} 

\definecolor{LightGreen}{HTML}{ccffcc}
\definecolor{Green}{HTML}{99ff99}
\definecolor{DarkGreen}{HTML}{66cc66}

\definecolor{LightRed}{HTML}{ffcccc}
\definecolor{Red}{HTML}{ff9999}
\definecolor{DarkRed}{HTML}{ff6666}

\definecolor{LightBlue}{HTML}{cce0ff}
\definecolor{Blue}{HTML}{99ccff}
\definecolor{DarkBlue}{HTML}{668cff}
\definecolor{LightTeal}{HTML}{B3FFFF}
\definecolor{MediumTeal}{HTML}{66FFFF}
\definecolor{DarkTeal}{HTML}{33CCCC}
\definecolor{LightYellow}{HTML}{FFFFCC}
\definecolor{MediumYellow}{HTML}{FFFF99}
\definecolor{DarkYellow}{HTML}{FFFF66}
\definecolor{lightgray}{gray}{0.9}

\definecolor{LightGreen}{HTML}{ccffcc}
\definecolor{Green}{HTML}{99ff99}
\definecolor{DarkGreen}{HTML}{66cc66}

\definecolor{LightRed}{HTML}{ffcccc}
\definecolor{Red}{HTML}{ff9999}
\definecolor{DarkRed}{HTML}{ff6666}

\definecolor{LightBlue}{HTML}{cce0ff}
\definecolor{Blue}{HTML}{99ccff}
\definecolor{DarkBlue}{HTML}{668cff}
\definecolor{LightTeal}{HTML}{B3FFFF}
\definecolor{MediumTeal}{HTML}{66FFFF}
\definecolor{DarkTeal}{HTML}{33CCCC}
\definecolor{LightYellow}{HTML}{FFFFCC}
\definecolor{MediumYellow}{HTML}{FFFF99}
\definecolor{DarkYellow}{HTML}{FFFF66}
\definecolor{lightgray}{gray}{0.9}

\newcommand{\plusvalue}[1]{\hspace{0.3em}\textcolor{darkgreen}{(+#1)}}
\newcommand{\minusvalue}[1]{\hspace{0.3em}\textcolor{black}{(-#1)}}
\definecolor{darkgreen}{rgb}{0.0, 0.5, 0.0}
\definecolor{MidGreen}{HTML}{AAFFAA}

\usepackage{xspace}
\def\ours{\textbf{MMEvol}\xspace}

\title{MMEvol: Empowering Multimodal Large Language Models with Evol-Instruct}

\author{Run Luo\textsuperscript{1,2}\thanks{Equal contribution. $\dag$ Jingkuan Song, Min Yang and Yongbin Li are corresponding authors.} \hspace{-1.0em}
\authorskip Haonan Zhang\textsuperscript{3}$^*$ \hspace{-1.5em}
\authorskip Longze Chen\textsuperscript{1,2}$^*$ \hspace{-1.5em}
\authorskip Ting-En Lin\textsuperscript{3}$^*$ \hspace{-1.5em}
\authorskip Xiong Liu\textsuperscript{3} \hspace{-1.5em} 
\authorskip Yuchuan Wu\textsuperscript{3} \hspace{-1.5em} \\ 
\textbf{Min Yang\textsuperscript{1,2}$^\dag$} \hspace{-1.5em}
\textbf{\authorskip Minzheng Wang\textsuperscript{2}} \hspace{-1.5em}
\textbf{\authorskip Pengpeng Zeng\textsuperscript{4}} \hspace{-1.5em}
\textbf{\authorskip Lianli Gao\textsuperscript{5}} \hspace{-1.5em}
\textbf{\authorskip Heng Tao Shen\textsuperscript{4}} 
\textbf{Yunshui Li\textsuperscript{1,2}}
\\
\textbf{Xiaobo Xia\textsuperscript{6}} 
\hspace{-1.5em}
\textbf{\authorskip Fei Huang\textsuperscript{3}} 
\hspace{-1.5em}
\textbf{\authorskip Jingkuan Song\textsuperscript{4}$^\dag$} \hspace{-1.5em}
\textbf{\authorskip Yongbin Li\textsuperscript{3}$^\dag$} \\[2mm]
\textsuperscript{1}Shenzhen Institute of Advanced Technology, Chinese Academy of Sciences \hspace{5.5mm} \\
\textsuperscript{2}University of Chinese Academy of Sciences \hspace{5.5mm} 
\textsuperscript{3}Alibaba Group \hspace{5.5mm} \\ \textsuperscript{4}Tongji University 
\hspace{5.5mm} \textsuperscript{5}Independent Researcher \hspace{5.5mm} \textsuperscript{6}The University of Sydney   \\
\texttt{\{r.luo, min.yang\}@siat.ac.cn} \\
\texttt{\{ting-en.lte, shuide.lyb\}@alibaba-inc.com} 
}

\iclrfinalcopy 
\begin{document}

\maketitle

\begin{abstract}

The development of Multimodal Large Language Models (MLLMs) has seen significant advancements with increasing demands in various fields (e.g., multimodal agents, embodied intelligence). While model-driven approaches attempt to enhance MLLMs capabilities through diverse architectures, the gains have become increasingly marginal. Conversely, data-driven methods, which scale up image-text instruction data, are more effective but face limited data diversity and complexity challenges. The absence of high-quality data constitutes a significant development barrier for MLLMs. To address the data quality bottleneck, we propose \ours, a novel multimodal instruction data evolution framework. This framework iteratively improve data quality through a refined combination of fine-grained perception, cognitive reasoning, and interaction evolution, generating a more complex and diverse image-text instruction dataset that empowers MLLMs with enhanced capabilities. Beginning with an initial set of instructions, SEED-163K, we utilize \ours to systematically broaden the diversity of instruction types, extend visual reasoning steps to improve cognitive reasoning abilities, and thoroughly explore fine-grained information within images to enhance visual understanding and robustness. To comprehensively evaluate the effectiveness of our approach, we conduct extensive qualitative analysis and quantitative experiments across 13 vision-language tasks. Compared to baseline models trained with the initial seed data, the results demonstrate that our method achieves an average accuracy improvement of 3.1 percentage points. Furthermore, our approach reaches state-of-the-art (SOTA) performance in nine tasks using significantly less data compared to state-of-the-art models. The project page is available at \url{https://mmevol.github.io/}
\end{abstract}

\section{Introduction}
\setlength{\epigraphwidth}{0.95\columnwidth}
\renewcommand{\epigraphflush}{center}
\renewcommand{\textflush}{flushepinormal}
\renewcommand{\epigraphsize}{\footnotesize}
\epigraph{\textcolor{black}{\textit{``The True Acquisition of Knowledge Lies in Grasping the Most Subtle Details.''}}}
{\textcolor{black}{\textit{Aristotle, circa 4th century BCE}}}
\vspace{-11pt}

Multimodal Large Language Models (MLLMs)~\citep{llava,llava-next,blip2,dreamllm,emu,instructblip,deem,cogcom} have seen rapid development over the past two years and have become the preferred approach for various vision-language tasks~\citep{ai2d,blink,mm-cot,miabench}. By aligning visual encoders~\citep{clip,sig-clip,eva-clip} with LLMs~\citep{llama,qwen,deepseek,yi,evol_survey}, and employing large-scale coarse-grained image-text pre-training~\citep{mmc4,laion-en,laion-400m} followed by small-scale instruction-tuning~\citep{allava,llava}, MLLMs have demonstrated impressive capabilities across numerous vision-language tasks and are widely applied in many domains (e.g., multimodal agents, embodied intelligence). Model-driven approaches~\citep{deem,llava-next,cambrian-10m,slime} aim to integrate knowledge from images and text more efficiently by designing different network architectures to improve MLLMs performance. However, their effectiveness is diminishing due to the lack of high-quality data, and redundant model designs struggle to realize their potential and push the boundaries of model intelligence. In contrast, data-driven methods~\citep{llava,allava,capsfusion,mminstruct,vila2,sharegpt4v}, while more effective, often produce data with limited diversity and complexity. The lack of high-quality data remains a significant obstacle in the development of MLLMs. Therefore, the need of developing automated methods that can generate more challenging and diverse instructional data at a relatively low cost is urgent for empowering MLLMs with enhanced capabilities.

\begin{figure}[t]
\begin{center}
\includegraphics[width=\linewidth]{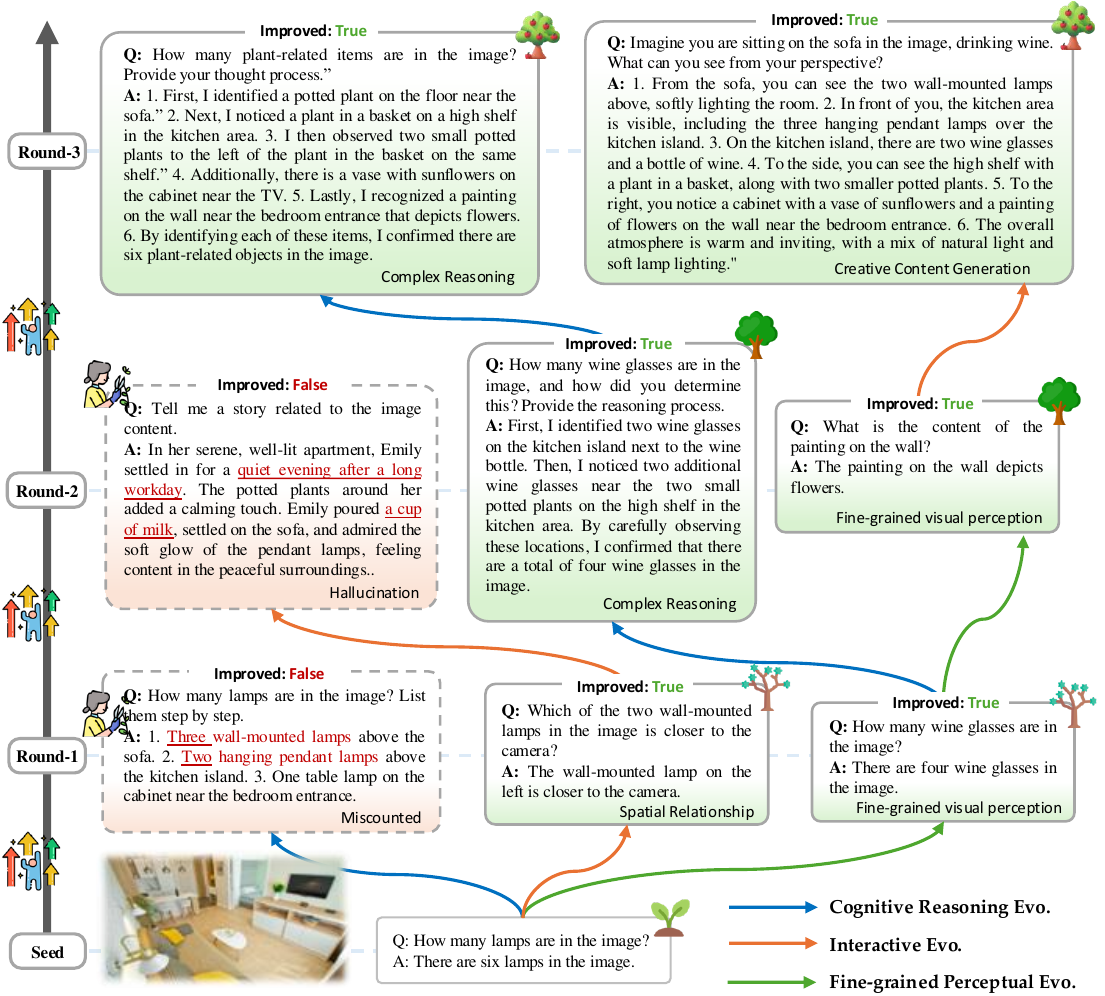}
\end{center}
\caption{\textbf{Overview of \ours.} 
Instruction evolution and instruction elimination synergistically collaborate through multiple rounds to enhance the diversity and complexity of instruction data.}
\label{fig:overview}
\vspace{-0.8cm}
\end{figure}

Analysis of existing data-driven methods for generating image-text instruction data reveals three common limitations: 1) \textbf{Limited instruction diversity}. Manually annotated instructions are constrained by the cognitive limitations of annotators, while model-generated instructions are limited by template presets, making it difficult to meet the diverse task requirements of the real world. This restricts the instruction-following ability of MLLMs. 2) \textbf{Limited instruction complexity}. Manual annotations often result in instructions of simple or moderate complexity, and automatically generated instructions tend to be brief and lacking in visual reasoning steps, which limits the model's ability to handle complex tasks. 3) \textbf{Insufficient alignment granularity}. Both manually and model-generated instructions primarily focus on common objects, neglecting rare or small objects, resulting in limited granularity in image-text alignment. This affects the model's visual perception robustness and resistance to hallucinations.

To address these limitations, we propose \ours, a novel method that utilizes advanced MLLMs for iterative evolution. This method automatically generates various types of open-domain instructions on a large scale, covering different difficulty levels to enhance the performance of MLLMs. Given that visual-language instruction data are constrained by visual content, the data generated through multiple iterations with Evol-Instruct~\citep{wizardlm,wizardmath,wizardcoder} tend to include simple restatements and data unrelated to visual content, making deep and broad evolution challenging. Therefore, we have made several adjustments to the evolution prompting process, ultimately developing an image-text instruction evolution paradigm. These adjustments include a more refined image-text instruction data paradigm and the definition of three evolution directions: fine-grained perception evolution, cognitive reasoning evolution, and interaction evolution. The \ours mechanism is summarized in \cref{fig:overview}, with each evolution cycle comprising two main steps: instruction evolution and instruction elimination. Instruction evolution randomly selects one of fine-grained perception evolution, cognitive reasoning evolution, or interaction evolution, upgrading simple instructions to more complex or diverse ones. Specifically, fine-grained perception evolution aims to leverage visual information in images to generate data with more detailed information; cognitive reasoning evolution prolongs the visual operation reasoning steps of instructions to increase their complexity; and interaction evolution aims to enhance instruction diversity by providing a wider variety of instruction forms. To account for occasional failures in evolved instructions, we use instruction elimination to filter out failed evolution. \ours repeats the instruction evolution and elimination processes multiple times to obtain a complex instruction dataset containing various instruction forms.

\begin{figure}[t]
    \centering
    \begin{minipage}{.34\textwidth}
        \includegraphics[height=125pt]{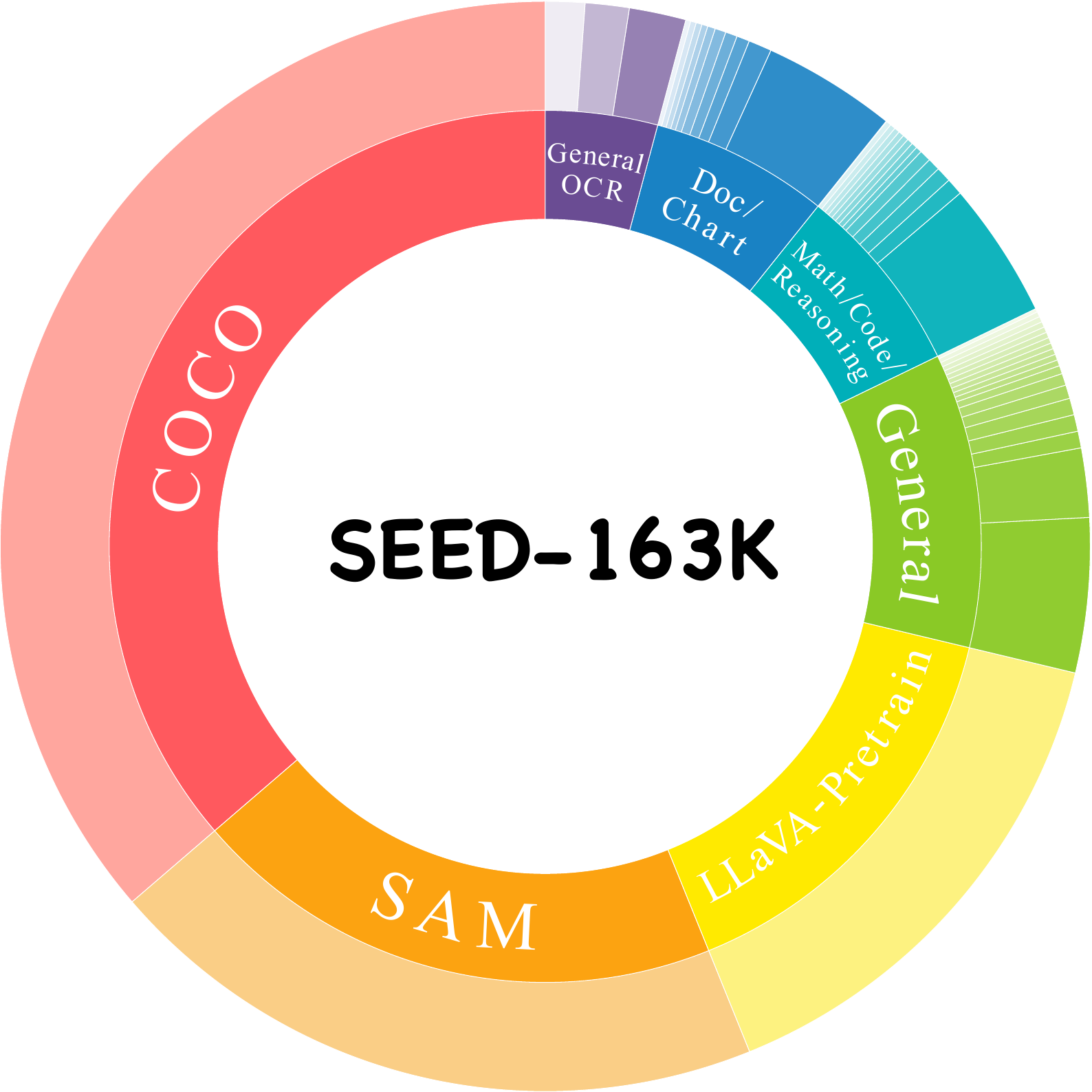}
    \end{minipage}%
    \hfill
    \begin{minipage}{.65\textwidth}
        \centering
        \renewcommand{\arraystretch}{1.1}
        \setlength\tabcolsep{1pt}
        \fontsize{6pt}{7.5pt}\selectfont
        \begin{tabular}{llll}

\makecell{\cellcolor[RGB]{255, 89, 94} \textcolor{white}{MS-COCO}} & \makecell{\cellcolor[RGB]{252, 163, 17} \textcolor{white}{SAM}} & \makecell{\cellcolor[RGB]{255, 234, 0} \textcolor{white}{LLaVA-Pretrain}} \\
\tikz[baseline=0.05em] \fill [color={rgb,255: red,255; green,166; blue,158}] (0,0) rectangle (0.75em,0.75em); COCO (96.6 K) & \tikz[baseline=0.05em] \fill [color={rgb,255: red,250; green,206; blue,134}] (0,0) rectangle (0.75em,0.75em); SAM (20.0 K)  & \tikz[baseline=0.05em] \fill [color={rgb,255: red,253; green,242; blue,128}] (0,0) rectangle (0.75em,0.75em); LLaVA-Pretrain (15.3 K) &    \\

\makecell{\cellcolor[RGB]{138, 201, 38} \textcolor{white}{General}}  & & & \\

\tikz[baseline=0.05em] \fill [color={rgb,255: red,144; green,204; blue,48}] (0,0) rectangle (0.75em,0.75em); CLEVER (4.6 K) & \tikz[baseline=0.05em] \fill [color={rgb,255: red,149; green,206; blue,59}] (0,0) rectangle (0.75em,0.75em); Q-Instruct-DB (2.0 K) & \tikz[baseline=0.05em] \fill [color={rgb,255: red,155; green,209; blue,69}] (0,0) rectangle (0.75em,0.75em); Web-Celebrity (0.5 K) &\tikz[baseline=0.05em] \fill [color={rgb,255: red,160; green,211; blue,79}] (0,0) rectangle (0.75em,0.75em); Web-Landmark (0.5 K) \\ 

\tikz[baseline=0.05em] \fill [color={rgb,255: red,166; green,214; blue,89}] (0,0) rectangle (0.75em,0.75em); Wikiart (0.5 K) & \tikz[baseline=0.05em] \fill [color={rgb,255: red,171; green,216; blue,100}] (0,0) rectangle (0.75em,0.75em); VizWiz (0.4 K) & \tikz[baseline=0.05em] \fill [color={rgb,255: red,177; green,219; blue,110}] (0,0) rectangle (0.75em,0.75em); LLaVAR (0.35 K) & \tikz[baseline=0.05em] \fill [color={rgb,255: red,182; green,222; blue,120}] (0,0) rectangle (0.75em,0.75em); TextbookQA (0.24 K)\\

\tikz[baseline=0.05em] \fill [color={rgb,255: red,188; green,224; blue,130}] (0,0) rectangle (0.75em,0.75em); ST\_VQA (0.18 K) & \tikz[baseline=0.05em] \fill [color={rgb,255: red,193; green,227; blue,141}] (0,0) rectangle (0.75em,0.75em); Hateful Memes (0.17 K) & \tikz[baseline=0.05em] \fill [color={rgb,255: red,199; green,229; blue,151}] (0,0) rectangle (0.75em,0.75em); IconQA (0.17 K) & \tikz[baseline=0.05em] \fill [color={rgb,255: red,204; green,232; blue,161}] (0,0) rectangle (0.75em,0.75em); GPT4V-Dataset (0.17 K) \\

\tikz[baseline=0.05em] \fill [color={rgb,255: red,210; green,235; blue,171}] (0,0) rectangle (0.75em,0.75em); Visual7w (0.17 K)  & \tikz[baseline=0.05em] \fill [color={rgb,255: red,215; green,237; blue,182}] (0,0) rectangle (0.75em,0.75em); Intergps (0.17 K) & \tikz[baseline=0.05em] \fill [color={rgb,255: red,221; green,240; blue,192}] (0,0) rectangle (0.75em,0.75em); Alfworld (0.17 K) & \tikz[baseline=0.05em] \fill [color={rgb,255: red,226; green,242; blue,202}] (0,0) rectangle (0.75em,0.75em); Data\_engine (0.17 K) \\

\tikz[baseline=0.05em] \fill [color={rgb,255: red,232; green,245; blue,212}] (0,0) rectangle (0.75em,0.75em); VisText (0.17 K) & \tikz[baseline=0.05em] \fill [color={rgb,255: red,237; green,247; blue,223}] (0,0) rectangle (0.75em,0.75em); FinQA  (0.15 K) & \tikz[baseline=0.05em] \fill [color={rgb,255: red,243; green,250; blue,233}] (0,0) rectangle (0.75em,0.75em); ScienceQA (0.10 K) & \\  

\makecell{\cellcolor[RGB]{0, 175, 185} \textcolor{white}{Math/Code/Reasoning}}  & & & \\

\tikz[baseline=0.05em] \fill [color={rgb,255: red,17; green,180; blue,189}] (0,0) rectangle (0.75em,0.75em); DVQA (4.0 K) & \tikz[baseline=0.05em] \fill [color={rgb,255: red,34; green,185; blue,194}] (0,0) rectangle (0.75em,0.75em); ChartQA (0.5 K) & \tikz[baseline=0.05em] \fill [color={rgb,255: red,51; green,190; blue,198}] (0,0) rectangle (0.75em,0.75em); DocVQA (0.45 K) & \tikz[baseline=0.05em] \fill [color={rgb,255: red,67; green,195; blue,203}] (0,0) rectangle (0.75em,0.75em); AI2D (0.37 K) \\

\tikz[baseline=0.05em] \fill [color={rgb,255: red,84; green,200; blue,207}] (0,0) rectangle (0.75em,0.75em); ArxivQA (0.34 K) & \tikz[baseline=0.05em] \fill [color={rgb,255: red,101; green,205; blue,212}] (0,0) rectangle (0.75em,0.75em); Chart2Text (0.17 K) & \tikz[baseline=0.05em] \fill [color={rgb,255: red,118; green,211; blue,216}] (0,0) rectangle (0.75em,0.75em); Robut\_WTQ (0.17 K) & \tikz[baseline=0.05em] \fill [color={rgb,255: red,135; green,216; blue,220}] (0,0) rectangle (0.75em,0.75em); Robut\_WikiSQL (0.17 K) \\

\tikz[baseline=0.05em] \fill [color={rgb,255: red,152; green,221; blue,225}] (0,0) rectangle (0.75em,0.75em); HiTab (0.17 K) & \tikz[baseline=0.05em] \fill [color={rgb,255: red,169; green,226; blue,229}] (0,0) rectangle (0.75em,0.75em); TQA (0.16 K) & \tikz[baseline=0.05em] \fill [color={rgb,255: red,185; green,231; blue,234}] (0,0) rectangle (0.75em,0.75em); VisualMRC (0.16 K) & \tikz[baseline=0.05em] \fill [color={rgb,255: red,202; green,236; blue,238}] (0,0) rectangle (0.75em,0.75em); InfoGraphic\_VQA (0.16 K) \\

 \tikz[baseline=0.05em] \fill [color={rgb,255: red,219; green,241; blue,243}] (0,0) rectangle (0.75em,0.75em); TaT\_QA (0.15 K) & \tikz[baseline=0.05em] \fill [color={rgb,255: red,236; green,246; blue,247}] (0,0) rectangle (0.75em,0.75em); Synthdog-En (0.01 K) &  & \\

\makecell{\cellcolor[RGB]{25, 130, 196} \textcolor{white}{Doc/Chart}}  & & & \\

\tikz[baseline=0.05em] \fill [color={rgb,255: red,46; green,141; blue,201}] (0,0) rectangle (0.75em,0.75em); Tabmwp  (4.0 K) & \tikz[baseline=0.05em] \fill [color={rgb,255: red,67; green,152; blue,207}] (0,0) rectangle (0.75em,0.75em); VG (0.7 K) & \tikz[baseline=0.05em] \fill [color={rgb,255: red,88; green,164; blue,212}] (0,0) rectangle (0.75em,0.75em); GQA (0.38 K) & \tikz[baseline=0.05em] \fill [color={rgb,255: red,109; green,175; blue,217}] (0,0) rectangle (0.75em,0.75em); Geo170k (0.35 K) \\

\tikz[baseline=0.05em] \fill [color={rgb,255: red,131; green,186; blue,223}] (0,0) rectangle (0.75em,0.75em); MathVision (0.32 K) & \tikz[baseline=0.05em] \fill [color={rgb,255: red,152; green,197; blue,228}] (0,0) rectangle (0.75em,0.75em); Design2Code (0.23 K) & \tikz[baseline=0.05em] \fill [color={rgb,255: red,173; green,208; blue,233}] (0,0) rectangle (0.75em,0.75em); Raven (0.18 K) & \tikz[baseline=0.05em] \fill [color={rgb,255: red,194; green,220; blue,238}] (0,0) rectangle (0.75em,0.75em); Websight (0.17 K)\\

\tikz[baseline=0.05em] \fill [color={rgb,255: red,215; green,231; blue,244}] (0,0) rectangle (0.75em,0.75em); Datikz (0.17 K) & \tikz[baseline=0.05em] \fill [color={rgb,255: red,236; green,242; blue,249}] (0,0) rectangle (0.75em,0.75em); GeomVerse (0.14 K) & &  \\

\makecell{\cellcolor[RGB]{106, 76, 147} \textcolor{white}{General OCR}}  & & & \\

\tikz[baseline=0.05em] \fill [color={rgb,255: red,150; green,129; blue,179}] (0,0) rectangle (0.75em,0.75em); TextVQA (0.7 K) & \tikz[baseline=0.05em] \fill [color={rgb,255: red,195; green,183; blue,211}] (0,0) rectangle (0.75em,0.75em); OCR\_VQA (0.3 K) & \tikz[baseline=0.05em] \fill [color={rgb,255: red,239; green,236; blue,243}] (0,0) rectangle (0.75em,0.75em); Rendered\_Text (0.17 K) \\

    \end{tabular}
    \end{minipage}
    \captionsetup{font={small}}
    \caption{\textbf{SEED-163K: 163K Curated Seed Instruction Tuning Dataset for Evol-Instruct.} \textbf{Left:} The inner circle shows the original distribution of SEED-163K. The outer circle shows the curated SEED-163K. \textbf{Right:} All the data sources in the SEED-163K dataset, as well as the ones filtered in data curation.}
    \label{fig:seed163k}
\vspace{-0.7cm}
\end{figure}

To validate the effectiveness of \ours, we perform three rounds of evolutionary iterations on 163K seed data, leading to 447K evolved samples. We fine-tuned the open-source LLaVA-NeXT~\citep{llava-next} model with these evolved data and compared it with other advanced methods across 13 vision-language benchmarks. Our method achieves state-of-the-art (SOTA) performance, demonstrating the effectiveness and efficiency of \ours. Additionally, we conduct detailed qualitative analysis and ablation experiments to showcase the contribution of each component of our method. We hope that the released evolutionary data and code will assist the community in understanding that using a small amount of high-quality image-text instruction data is far more critical than training MLLMs with large-scale low-quality image-text instruction data.

Our main contributions can be summarized as follows:
\begin{itemize}

\item A image-text instruction evolution framework, \ours, is designed to leverage advanced MLLMs, automating the generation of open-domain image-text instruction data across varying difficulty levels to enhance the diversity and complexity of existing datasets.
\item By utilizing instruction evolution data, a high-quality data recipe is composed, and the evolved data will be released to advance the capabilities of other open-source MLLMs further.
\item We train an MLLM using this high-quality data recipe, achieving superior performance in various downstream visual-language tasks compared to other fully open-source methods.
\item The effectiveness and efficiency of the proposed approach are validated through extensive qualitative and quantitative analyses.
\end{itemize}

\begin{figure}[h]
\begin{center}
\includegraphics[width=\linewidth]{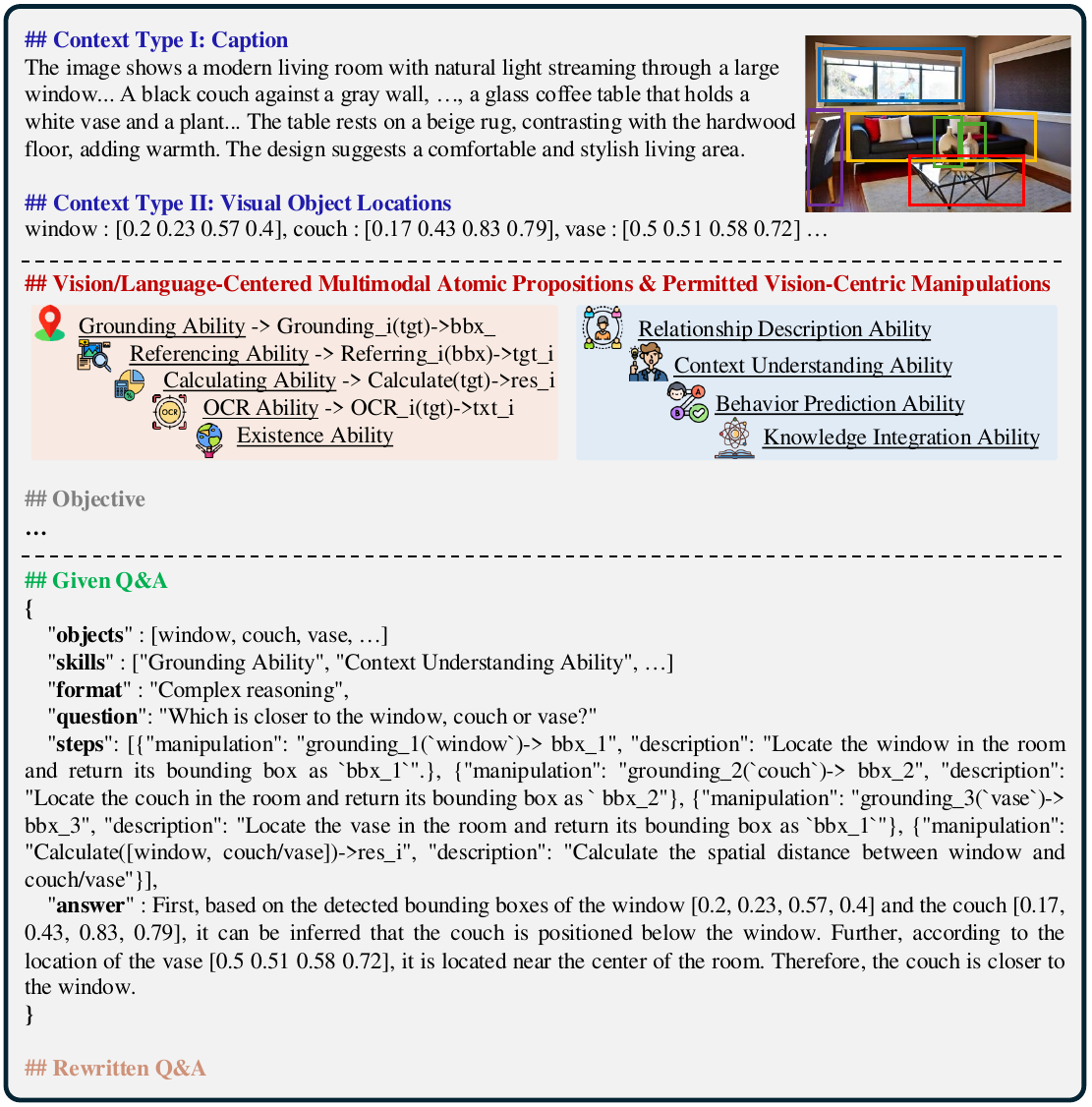}
\end{center}
\caption{\textbf{Prompt Head of \ours}. The top block showcases the contexts such as caption and visual object locations, and the middle block demonstrates vision/la
nguage-centered atomic propositions and evolution objective (described later). Additionally, we endow vision capabilities with pseudo-function calls to enhance visual reasoning during evolutionary processes. Finally, the bottom block further elucidates the organized seed sample, which is subsequently sent to the MLLM for rewriting.}
\label{fig:prompt_template}
\vspace{-0.7cm}
\end{figure}

\section{Method}
In this section, we first introduce the curation of seed instruction data and then elaborate on the methodological details of \ours. Due to the space limitation, we simplify the seed data curation process and prompt templates. More details can be found in the \cref{sec:add_prompt_details}.

\subsection{Seed Data Curation}

The seed instruction data are curated from LLaVA-Instruct~\citep{llava} and ShareGPT4V ~\citep{sharegpt4v} datasets, supplemented with additional scientific and chart data sampled from Cambrain-1 ~\citep{cambrian-10m}. This process involved careful selection and refinement to ensure the quality and diversity of the instructions. For instructions with only captions, we use the OpenAI GPT-4o mini API to generate seed instruction data. Ultimately, after merging and filtering, we obtained a comprehensive dataset consisting of 163K instruction samples with unique images, which serve as the foundation for our subsequent Evol-Instruct. The seed data mixture is shown in \cref{fig:seed163k}.
Please refer to \cref{sec:seed_details} for more details.

\subsection{Methodological Details}

\begin{figure}[t]
\begin{center}
\includegraphics[width=\linewidth]{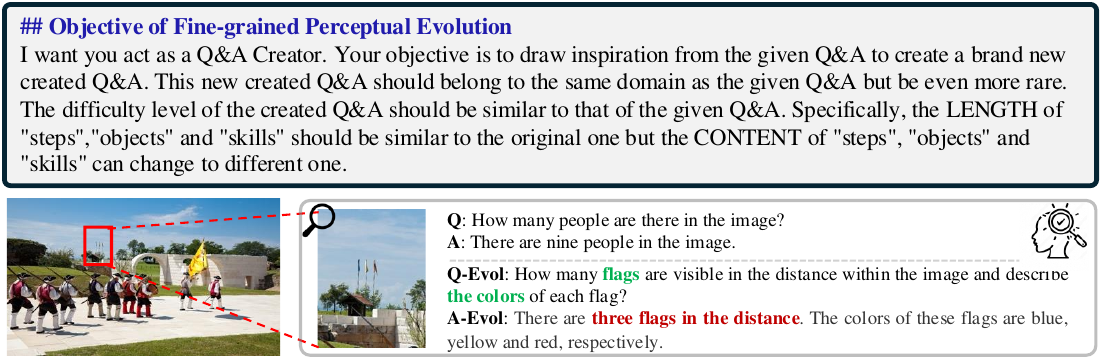}

\end{center}
\caption{\textbf{Fine-grained perceptual evolution prompt and example.} Fine-grained perceptual evolution can generate samples with more detailed visual information, enhancing data diversity, which are marked with different colors for better visualization.}
\label{fig:fine_perceptual_evol}
\vspace{-0.7cm}

\end{figure}

\begin{figure}[h]
\begin{center}
\includegraphics[width=\linewidth]{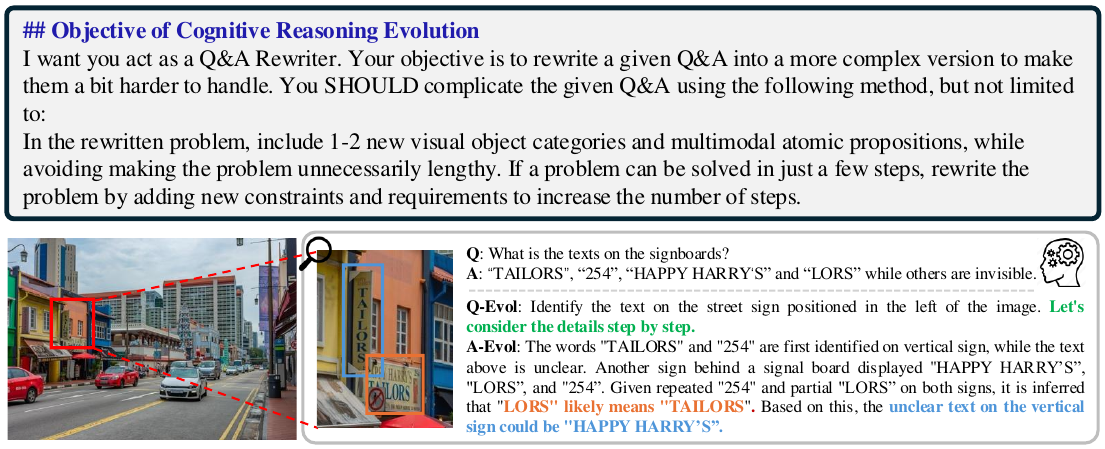}

\end{center}
\caption{\textbf{Cognitive reasoning evolution prompt template and example.} Cognitive reasoning evolution can endow instruction data with a longer visual reasoning chain, increasing the complexity of the data. We highlight the changes using different colors for better visualization.}
\label{fig:reasoning_evol}
\vspace{-0.4cm}
\end{figure}

The evolution of image-text instruction data is constrained by visual information, requiring evolved instruction data to be relevant to image content to avoid hallucinations. This makes the diversity evolution of image-text instructions particularly challenging. Additionally, the complexity evolution process of image-text instruction data often results in shallow reasoning phenomena, with MLLMs struggling to provide complex answers. As shown in \cref{fig:overview}, to address these issues and improve the success rate of evolution, we include carefully designed domains such as visual objects, atomic capabilities, visual manipulations, and instruction formats to standardize each instruction data format. The visual object domain includes visual objects in the images involved in the instruction data, implicitly constraining the evolution data and reducing visual hallucinations. We also summarize nine types of atomic capabilities involved in image-text instruction data to populate the atomic capability domain, aiming to enhance data diversity. Specifically, this includes five vision-centric capabilities: localization, reference, computation, optical character recognition (OCR), and existence judgment, and four language-centric capabilities: relation description, scene understanding, behavior prediction, and world knowledge association. The visual manipulation domain includes visual manipulation chains for problem-solving, where each step of the visual manipulation is based on vision-centric atomic capabilities, explicitly defining the visual reasoning process to mitigate shallow reasoning. The instruction format domain specifies the interaction types of the instruction data. These adaptations enhance the diversity and complexity of image-text instruction data and improve the success rate of evolution.

\myparagraph{Fine-grained Perceptual Evolution.} The goal of fine-grained perceptual evolution is to maximize the extraction of available visual information from images, especially overlooked non-primary visual objects. We observe that most instruction data tend to construct questions involving primary objects in images while neglecting less frequent non-primary objects. This results in a lack of instructions related to long-tail distribution objects. Training with such data can lead to visual hallucinations and poor generalization and robustness. Fine-grained perceptual evolution generates questions involving new visual objects, uncovering usable and often overlooked visual information. The evolutionary prompt template and process are shown in \cref{fig:fine_perceptual_evol}.

\begin{figure}[t]
\begin{center}
\includegraphics[width=\linewidth]{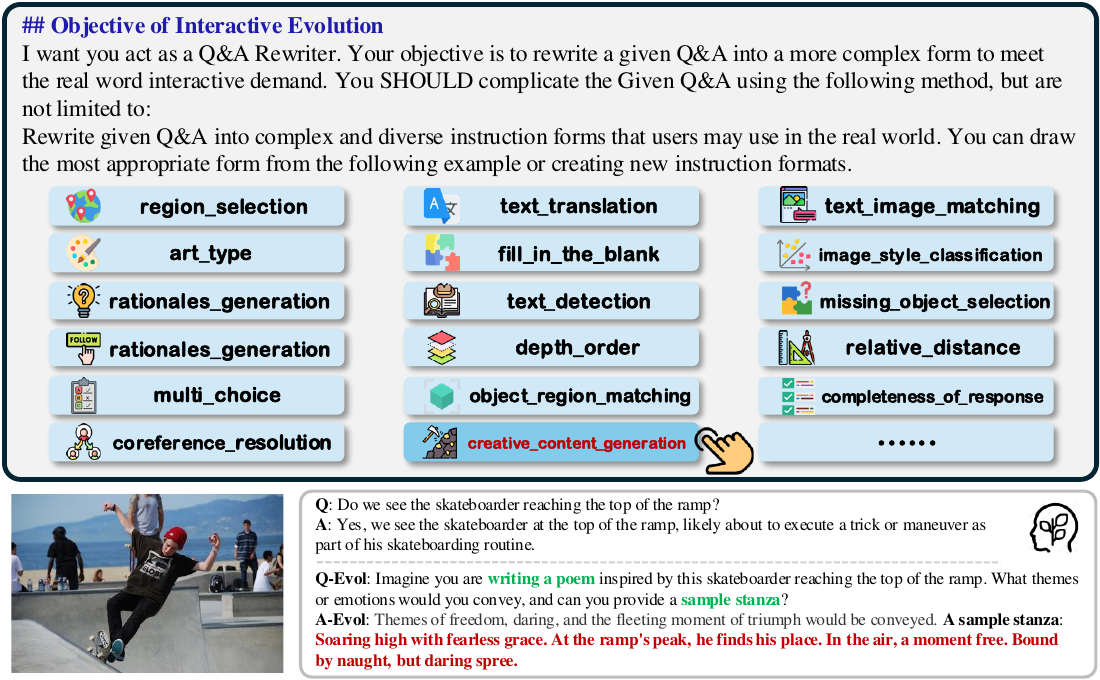}

\end{center}
\caption{\textbf{Interactive evolution prompt template and example.} Interactive evolution can automatically generate various types of non-predefined instruction formats, significantly enhancing the diversity of the data. The differences are highlighted using distinct colors for better visualization.}
\label{fig:interactive_evol}
\vspace{-0.7cm}
\end{figure}

\begin{figure}[h]
\begin{center}
\includegraphics[width=\linewidth]{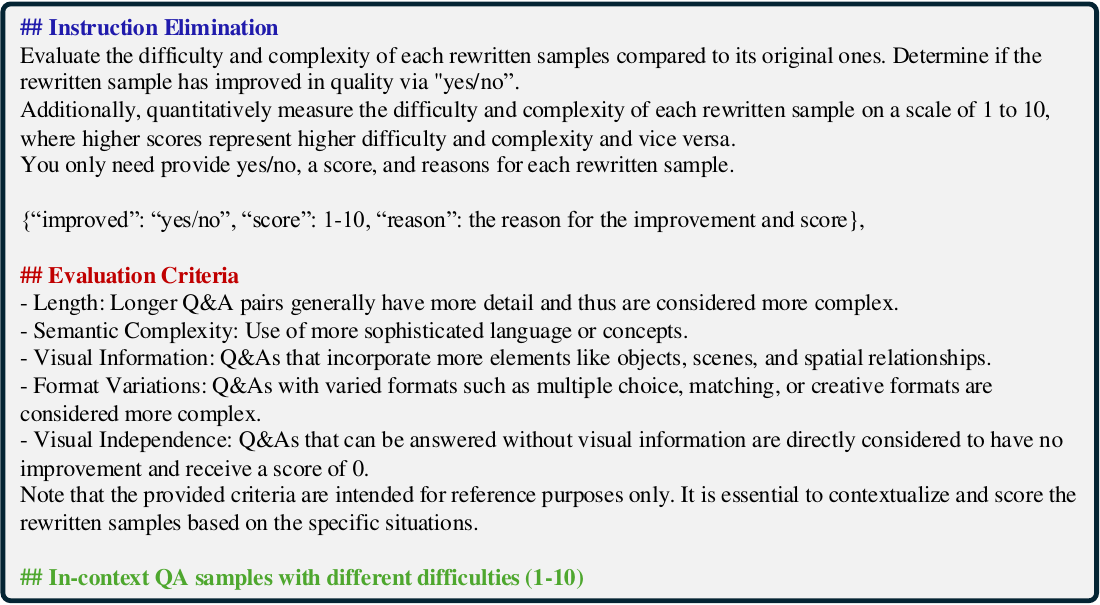}

\end{center}
\caption{\textbf{Instruction elimination prompt template.} Instruction elimination is used to calculate the evolutionary gain and complexity level of the instruction data. We filter out harmful data that failed to evolve based on the evolutionary gain.}
\label{fig:instruction_elimination}
\end{figure}

\myparagraph{Cognitive Reasoning Evolution.} Reasoning ability is one of the key capabilities of multi-modal large language models. However, most existing instruction data, such as LLaVA-Instruct~\citep{llava}, consists of simple question-and-answer pairs that lack detailed reasoning processes, making it difficult for trained models to accomplish complex tasks requiring reasoning capabilities, such as multi-modal agents and visual reasoning. We introduce the concept of a visual manipulation chain, abstracting four vision-centric reasoning capabilities into four visual operation functions described in text. By generating the necessary visual reasoning steps to solve problems, we define the complexity of the instruction data. During the cognitive reasoning evolution process, we evolve new instruction data by increasing the visual reasoning steps in the data to obtain more complex data. The evolutionary prompt template and process are shown in \cref{fig:reasoning_evol}.

\myparagraph{Interactive Evolution.} Existing models generate instruction data in very few forms. For example, LLaVA-Instruct provides only dialogue-based question-answering, complex reasoning, and global description tasks. Handcrafted instruction data, such as ALLaVA~\citep{allava}, are limited by annotators' experience, making it challenging to design various task forms. Models trained with such data often struggle to follow complex and diverse user-specified instructions or goals, limiting their practicality and applicability in real-world scenarios. To evolve instruction data with rich task forms and provide a good interaction experience, we design interactive evolution to generate instruction data with diverse task forms automatically. The evolutionary prompt template and process are demonstrated in \cref{fig:interactive_evol}.

\textbf{Instruction Elimination}. After each round of evolution, we score the evolved instruction data on multiple dimensions to assess the success of the evolution. We retain instruction data with evolutionary gains and discard those with failed evolution. The evolutionary elimination prompt template and process are shown in \cref{fig:instruction_elimination}.

\section{Experiments}

\subsection{Benchmarks}
To comprehensively evaluate the effectiveness of our evolutionary method, we select 13 benchmarks, with their sources and tested skills illustrated in \cref{tab:benchmark}. MIA~\citep{miabench} is an open-domain instruction-following benchmark that thoroughly tests the model’s instruction-following abilities using extensive instruction data. MM-Self-Instruct~\citep{mm-cot} is a novel visual reasoning benchmark that focuses on the model’s visual perception capabilities and performs common visual reasoning tasks encountered in daily life.

\subsection{Implementation Details}

\textbf{Data.} During the pre-training phase, we use LLaVA-Pretrain-595K~\citep{llava} for image-text alignment training. In ablation experiment settings, we fine-tune using both seed data and evolved data separately to ensure a fair comparison and validate the benefits of \ours. In SOTA setting experiments, we fine-tune using evolved instruction data combined with other publicly available datasets sampled from Cambrain-1~\citep{cambrian-10m} and compare it with other methods. Additional details on training data recipes can be found in the \cref{sec:recipe_details}.

\textbf{Model.} We follow the architecture from LLaVA-NeXT, where a multimodal large model consists of three key components: an LLM for next token prediction, a visual encoder for extracting visual features, and an image-text projector to align the visual and text modalities. We use Llama3-8B-Instruct~\citep{llama} for ablation experiments. For comparisons with other methods, we switch to our previous SOTA settings with Llama3-8B-Instruct and Qwen2-7B-Instruct~\citep{qwen}. We adapt CLIP-ViT-L~\citep{clip} for the visual encoder and use simple linear layers to bridge the image and text modalities.

\textbf{Training Strategies.} We conduct \ours
training following widely used two-stage settings. Vision-Language Pre-training and Visual Instruction-tuning. The language models and ViT are separately pre-trained, while the projector is randomly initialized. To initially align the feature space 
between the visual and text modalities, we utilize the aligned dataset. Finally, we perform instruction tuning of the pre-trained model on visual language instruction datasets. Our experiments are conducted with 8$\times$A100 GPUs and a global batch
size of 128. We employ AdamW optimizer~\citep{adamw} with learning rates $5 \times 10^{-5}$ and  $2 \times 10^{-5}$ for aforementioned two stages respectively. Each stage is trained with one epoch with a 3\% warmup
strategy. Please refer to the \cref{sec:recipe_details} for more details.

\subsection{Qualitative Analysis}

\begin{wrapfigure}{R}{0.54\linewidth}
   \vspace{-0.7cm}
    \centering
    \begin{subfigure}{0.49\linewidth}
        \centering
        \includegraphics[width=\linewidth]{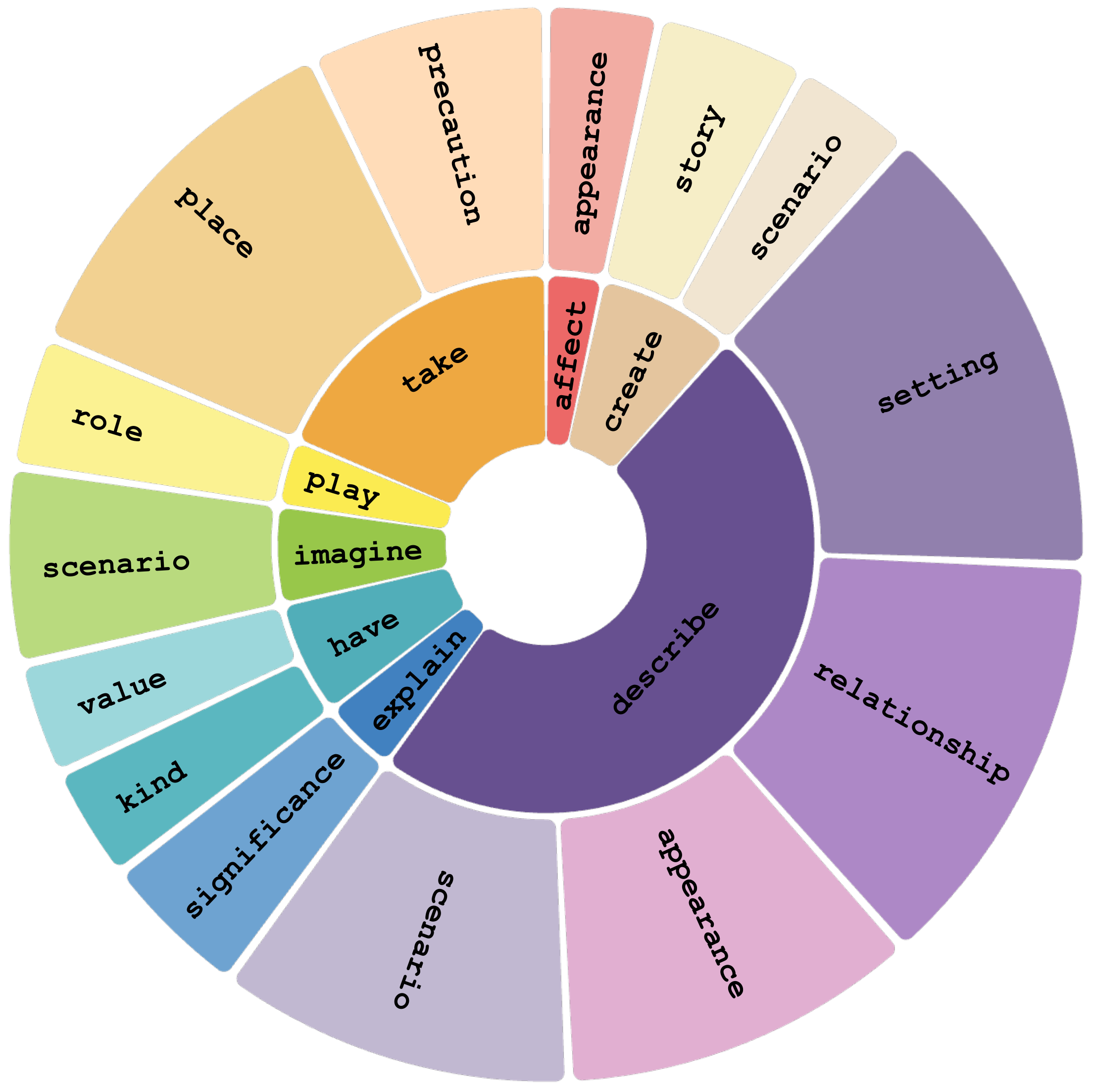}
        \caption{Seed data}
        \label{fig:before}
    \end{subfigure}
    \hfill
    \begin{subfigure}{0.49\linewidth}
        \centering
        \includegraphics[width=\linewidth]{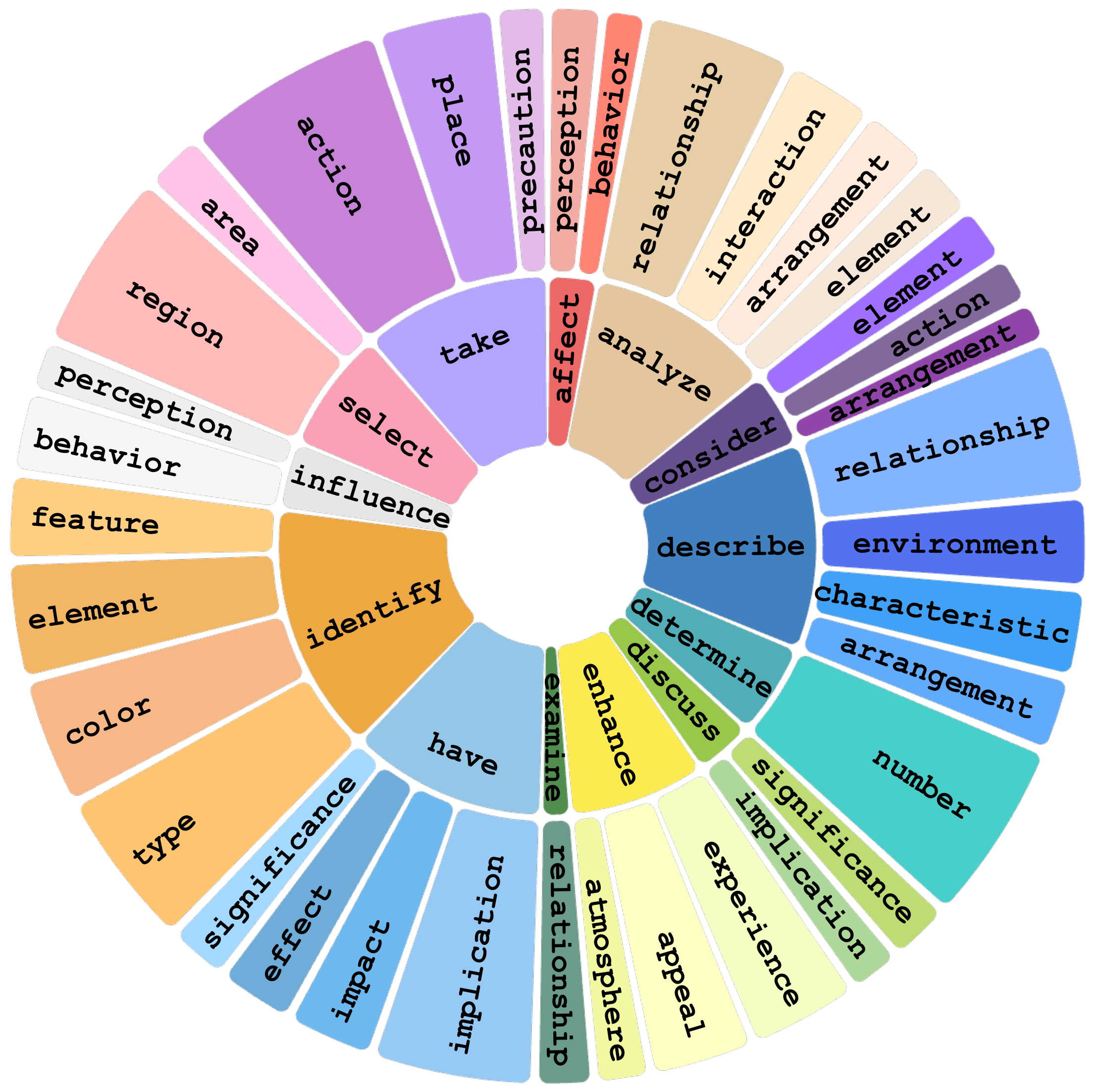}
        \caption{Evolved data}
        \label{fig:after}
    \end{subfigure}
    \caption{The root verbs (inner circle) and their top noun objects (outer circle) of the seed data in (a) and  the evolved data in (b).}
    \label{fig:instruct_dis}
    \vspace{-0.5cm}
\end{wrapfigure}

\begin{figure}[ht]
    \centering
    \begin{subfigure}{0.3\columnwidth}
        \centering
        \includegraphics[width=\columnwidth]  {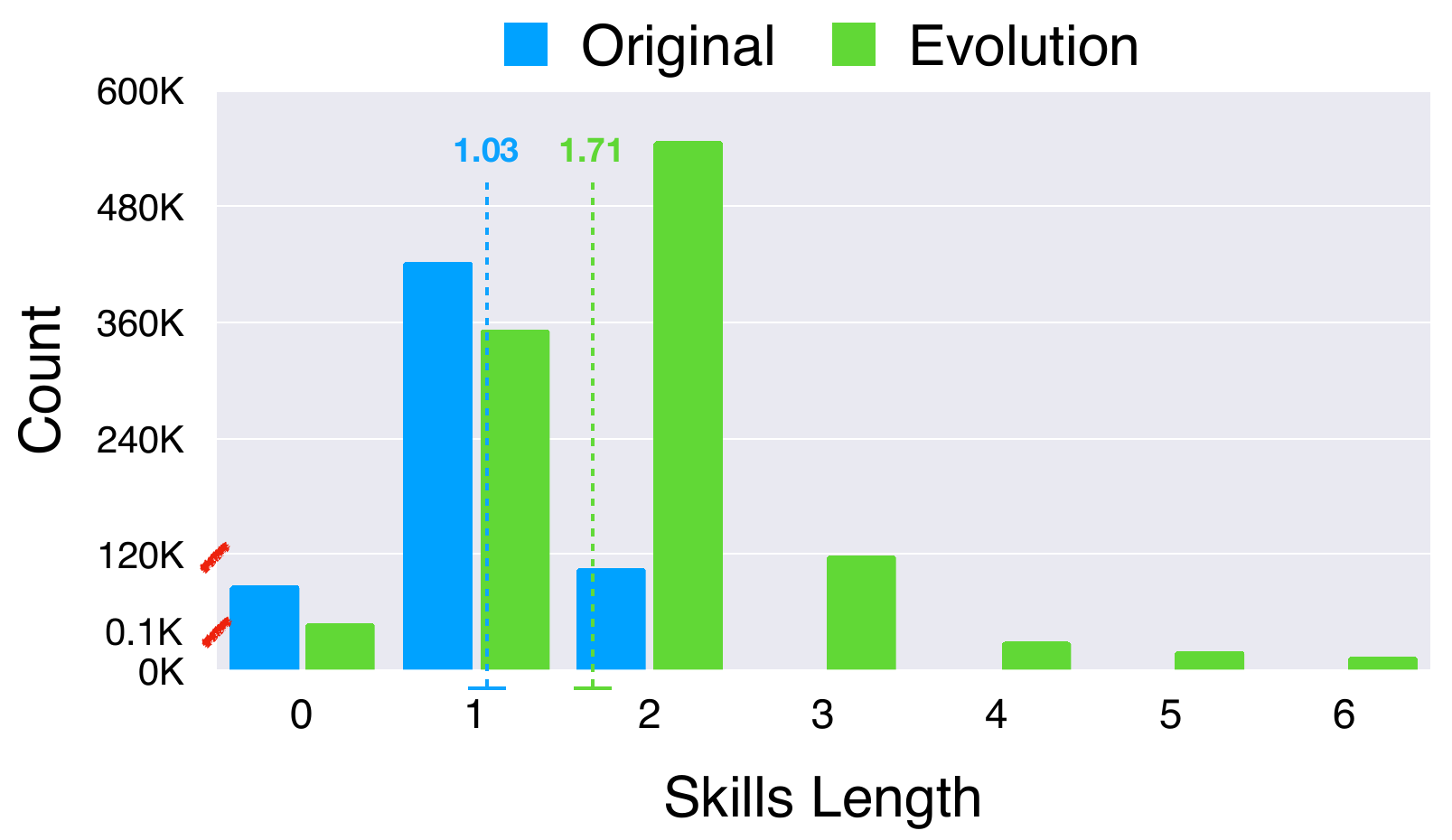}
        \caption{}
        \label{fig:vis-skill}
    \end{subfigure}%
        \begin{subfigure}{0.3\columnwidth}
        \centering
        \includegraphics[width=\columnwidth]{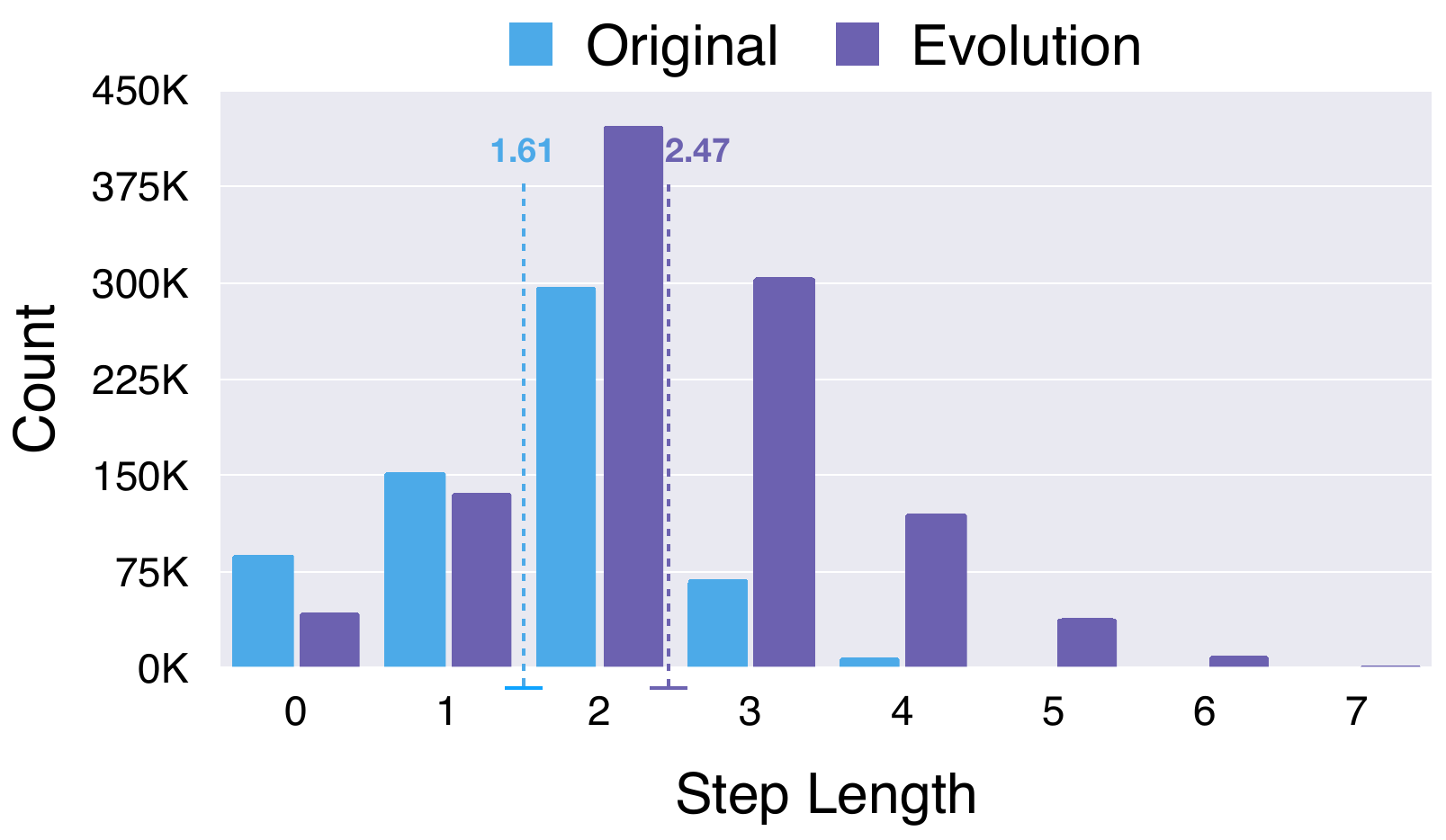}
        \caption{}
        \label{fig:vis-step}
    \end{subfigure}%
    \begin{subfigure}{0.32\columnwidth}
        \centering
        \includegraphics[width=\columnwidth]{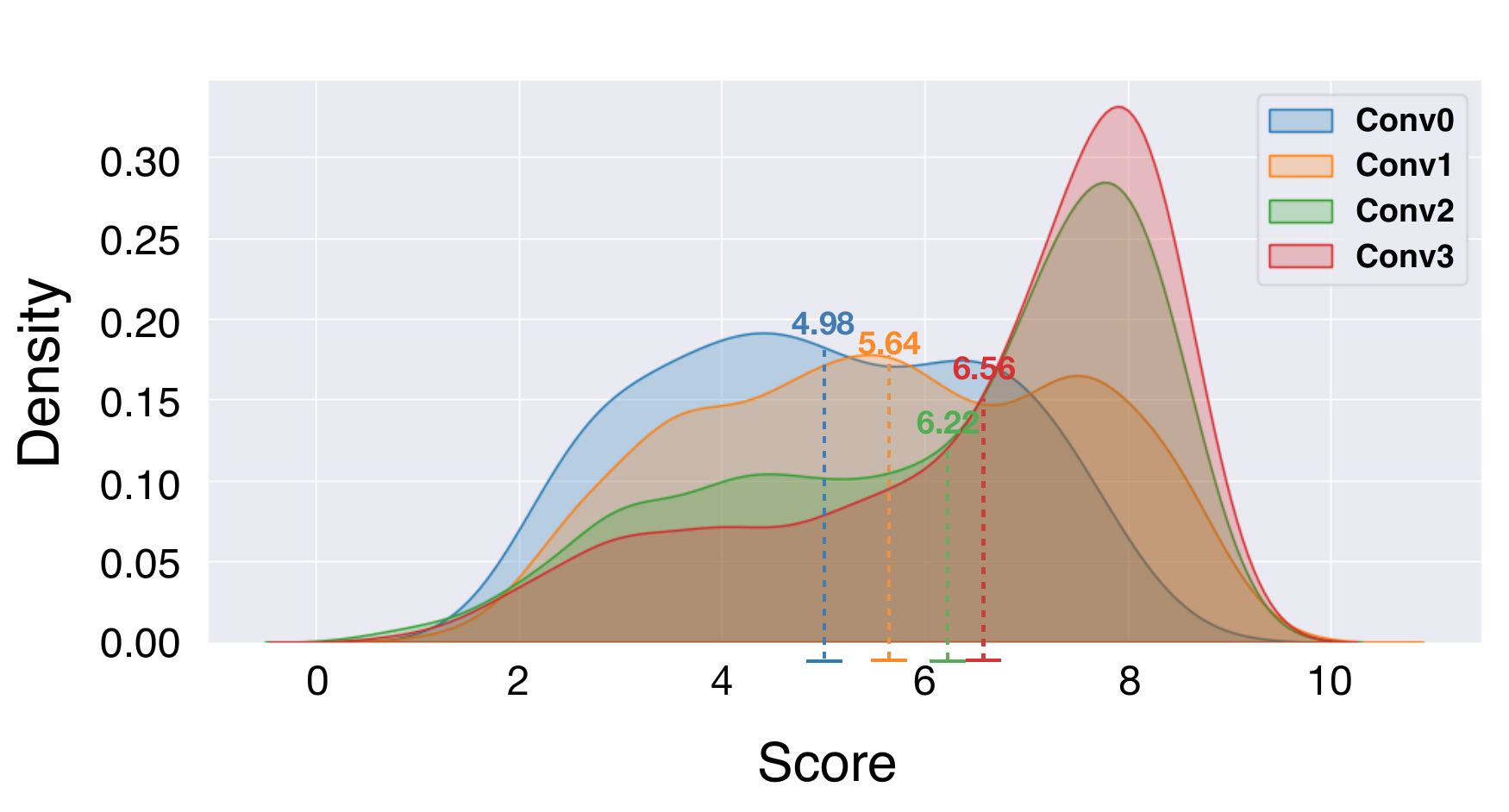}
        \caption{}
        \label{fig:vis-score}
    \end{subfigure}
    \caption{(a) The skills length distribution between the seed data and our evolved data; (b) The reasoning steps length distribution between the seed data and our evolved data; (c) The difficulty and complexity level distribution between the seed data and our evolved data.}
    \label{fig:vis-complexity}
\end{figure}

We randomly sample 30K data points from the seed data and conduct qualitative analysis on the instruction data before and after evolution. As shown in \cref{fig:vis-complexity}, the evolved data is notably more complex. Specifically, each evolved instruction involves 0.68 more atomic abilities in \cref{fig:vis-skill} and has an average visual operation chain reasoning length of 0.86 longer compared with pre-evolution in \cref{fig:vis-step}. As we can see from \cref{fig:vis-score}, the average difficulty score of each evolution round increases progressively, demonstrating the effectiveness of cognitive reasoning evolution in increasing instruction data complexity.

\begin{figure}[ht]
\begin{center}
\includegraphics[width=\linewidth]{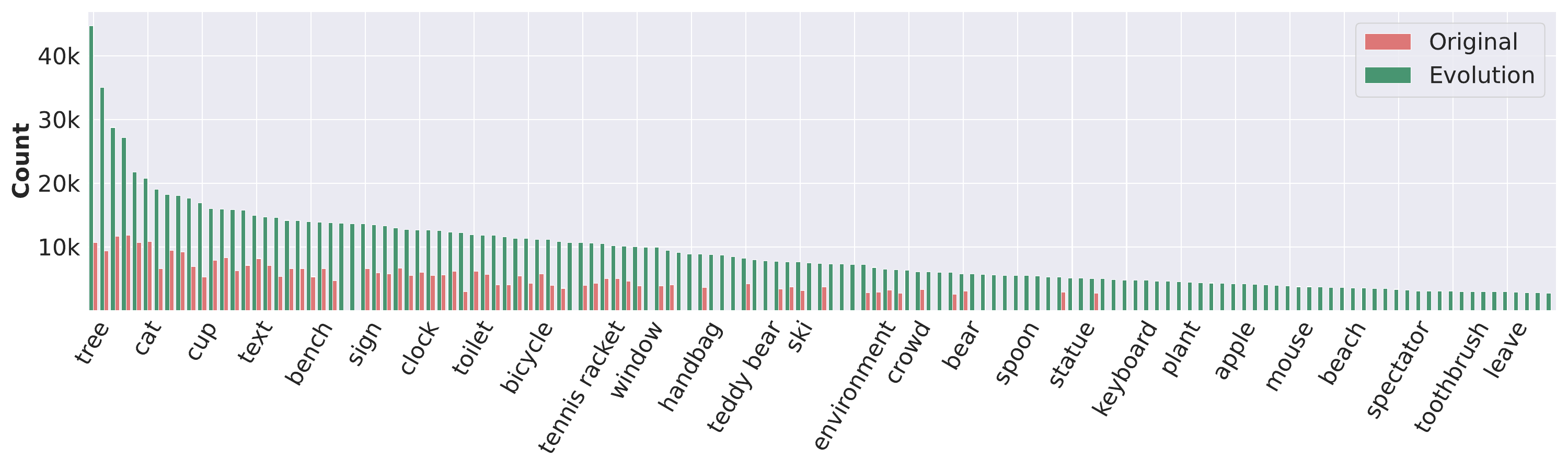}
\end{center}
\caption{The long-tail distribution of 200 visual objects between seed and evolved data. \ours significantly improves the long-tail distribution of visual objects in the seed data, providing more fine-grained visual information, thereby boosting the model's generalization ability and robustness against hallucinations.}
\label{fig:dis_objects}
\vspace{-0.4cm}
\end{figure}

\label{sec:analysis}
We identify the verb-noun structures in the generated instructions to study the types of instructions generated and the diversity of evolved data. We use the Berkeley Neural Parser~\citep{parser-1,parser-2} to parse the instructions, extracting the verb closest to the root and its first direct noun object. \cref{fig:instruct_dis} plots the root verbs and their direct noun objects with quantities exceeding 2K. We observe that the evolved data significantly enhances instruction diversity compared to pre-evolution, with diverse intents and textual formats in the evolved instructions. Furthermore, we conduct a long-tail distribution visualization analysis of the visual object domain in the instruction data before and after evolution to verify the effectiveness of fine-grained perceptual evolution. \cref{fig:dis_objects} shows that fine-grained perceptual evolution greatly improves the distribution of visual objects in the long tail, maximizing the extraction of usable visual information from images, refining the image-text alignment granularity in the instruction data, enhancing data diversity, which improves model generalization and reduces visual hallucinations.

\subsection{Ablation Study}
\label{sec:ablation_study}
We conduct ablation studies on seven vision-language benchmarks to explore the effects of instruction evolution and elimination. As shown in  \cref{tab:ablation}, 
different evolution process can be orthogonally superimposed on each other to continuously enhances data diversity and complexity. leading to an average performance gain of 3.8 points across multiple vision-language benchmarks. However, the absence of instruction elimination introduces harmful data from failed evolutions, which inevitably reduces the model’s resistance to hallucinations by 1.2 points on POPE ~\citep{pope}. When both instruction evolution and instruction elimination are employed, instruction elimination filters out harmful data from failed evolutions, further improving the quality and density of evolved data and enhancing the model’s performance by 0.9 points on average, particularly improving resistance to hallucinations by 1.7 points, which aligns with our qualitative analysis results in \cref{sec:analysis}. 

In addition, we conduct an ablation study starting from 1k seed data, focusing on the ratios of three evolutionary directions. The results, shown in \cref{tab:ablation_ratio}, indicate that equal probabilities for the three directions yield the best average performance, confirming their importance in enhancing the complexity and diversity of augmented data. To demonstrate the effectiveness of our prompt design with explicit complexity and diversity metrics, we replace the evolved prompts with their simplest basic versions and conduct experiments on the 1k data. As shown in \cref{tab:ablation_prompt}, measurable complexity and diversity metrics significantly improve the quality of data evolution. Please refer to \cref{sec:add_ablation} for more ablation study.

\begin{table}[ht]
\centering
\caption{\textbf{Ablation study on instruction evolution and instruction elimination (6K).} The application of instruction evolution alone enhances the complexity and diversity of the data, whereas the integration of instruction elimination further refines data quality, markedly reducing the occurrence of visual hallucinations.}
\setlength{\tabcolsep}{4pt}
\renewcommand{\arraystretch}{1.2}
\resizebox{\textwidth}{!}{%
\begin{tabular}{llllcccccccc}
    \toprule
        \textbf{FP-Evol} & \textbf{I-Evol} & \textbf{CR-Evol} & \textbf{I-Elim} &\textbf{\small{MMStar}} & \textbf{\small{MathVista$\rm^{M}$}} & \textbf{\small{POPE}} & \textbf{\small{AI2D}} & \textbf{\small{MME$\rm^{C}$}} &  \textbf{\small{MMMU$\rm^{V}$}}  & \textbf{\small{RWQA}} & \textbf{\small{AVG.}}\\
        \midrule[0.4pt]
        \fmark& \fmark & \fmark & \fmark & 36.5 & 25.3 & 84.8 & 53.9 & 31.5 & 32.3 & 43.5 & 44.0 \\
         \cmark & \fmark & \fmark & \fmark & 37.3 {\textbf{\tiny{\plusvalue{0.8}}}} & 25.6 {\textbf{\tiny{\plusvalue{0.3}}}} & 85.0 {\textbf{\tiny{\plusvalue{0.2}}}} & 54.2
         {\textbf{\tiny{\plusvalue{0.3}}}} & 33.0 {\textbf{\tiny{\plusvalue{1.5}}}} & 32.5
         {\textbf{\tiny{\plusvalue{0.2}}}} & 46.7
         {\textbf{\tiny{\plusvalue{3.2}}}}
         & 44.9 {\textbf{\tiny{\plusvalue{0.9}}}} \\
        \cmark & \cmark & \fmark & \fmark & 38.2 {\textbf{\tiny{\plusvalue{1.7}}}} & 26.2 {\textbf{\tiny{\plusvalue{0.9}}}} & 83.8 {\textbf{\tiny{\minusvalue{1.0}}}} & 54.5 {\textbf{\tiny{\plusvalue{0.6}}}} & 35.6
        {\textbf{\tiny{\plusvalue{4.1}}}} & 32.9
        {\textbf{\tiny{\plusvalue{0.6}}}} & 48.9
        {\textbf{\tiny{\plusvalue{5.4}}}} & 45.8  {\textbf{\tiny{\plusvalue{1.8}}}}\\
        \cmark & \cmark & \cmark & \fmark & 38.9 {\textbf{\tiny{\plusvalue{3.4}}}}& 27.3 {\textbf{\tiny{\plusvalue{3.0}}}}& 83.6 {\textbf{\tiny{\minusvalue{1.2}}}} & 54.7 {\textbf{\tiny{\plusvalue{0.8}}}} & \textbf{40.1}
        {\textbf{\tiny{\plusvalue{8.6}}}} & 34.4
        {\textbf{\tiny{\plusvalue{0.9}}}}& 54.4
        {\textbf{\tiny{\plusvalue{10.9}}}}& 47.6 {\textbf{\tiny{\plusvalue{3.8}}}} \\
        \cmark & \cmark & \cmark & \cmark & \textbf{40.3} {\textbf{\tiny{\plusvalue{3.8}}}} & \textbf{28.6} {\textbf{\tiny{\plusvalue{3.6}}}} & \textbf{86.5} {\textbf{\tiny{\plusvalue{1.7}}}} & \textbf{55.2} {\textbf{\tiny{\plusvalue{1.3}}}} & 39.9 {\textbf{\tiny{\plusvalue{8.4}}}} & \textbf{35.3} {\textbf{\tiny{\plusvalue{3.0}}}} & \textbf{55.3}  {\textbf{\tiny{\plusvalue{11.8}}}} & \textbf{48.7} {\textbf{\tiny{\plusvalue{4.7}}}}\\
    \bottomrule
    \end{tabular}
}
\vspace{-1pt}
\label{tab:ablation}
\vspace{-2mm}
\end{table}

\begin{table}[ht]
    \centering
    \caption{\textbf{Ablation study on instruction evolution ratio.} The highest average performance can be achieved when the ratios of the three evolutionary directions are equal, thereby demonstrating that all three directions are equally important for the diversity and complexity of the evolutionary instruction data.}
    \setlength{\tabcolsep}{4pt}
    \renewcommand{\arraystretch}{1.2}
    \resizebox{\textwidth}{!}{%
    \begin{tabular}{llllcccccccc}
        \toprule
            \textbf{FP-Evol} & \textbf{I-Evol} & \textbf{CR-Evol} & \textbf{I-Elim} &\textbf{\small{MMStar}} & \textbf{\small{MathVista$\rm^{M}$}} & \textbf{\small{MME$\rm^{C}$}} & \textbf{\small{AI2D}} & \textbf{\small{HallBench}} &  \textbf{\small{MMMU$\rm^{V}$}}  & \textbf{\small{RWQA}} & \textbf{\small{AVG.}}\\
            \midrule[0.4pt]
            2/3 & 1/6 & 1/6 & \cmark & 36.9 & 26.3 & 31.0 & 54.0 & \textbf{44.8} & 34.4 & 51.4 & 39.8 \\
            1/6 & 2/3 & 1/6 & \cmark & 34.3 & 25.4 & 29.2 & 53.2 & 43.5 & \textbf{35.8} & 52.6 & 39.2 \\
            1/6 & 1/6 & 2/3 & \cmark & 36.3 & \textbf{26.7} & \textbf{32.5} & 54.3 & 44.0 & 35.2 & 51.1 & 40.0 \\
            1/3 & 1/3 & 1/3 & \cmark & \textbf{37.9} & 26.1 & 31.3 & \textbf{55.1} & 43.8 & \textbf{35.8} & \textbf{53.2} & \textbf{40.5} \\
        \bottomrule
        \end{tabular}
    }
    \vspace{-1pt}
    \label{tab:ablation_ratio}
    \vspace{-2mm}
\end{table}

\begin{table}[ht]
    \centering
    \caption{\textbf{Ablation study on prompt version.} Utilizing our meticulously designed prompts significantly enhances the diversity and complexity of the data, thereby making the evolutionary process more efficient. The symbol \cmark
 indicates that during the evolutionary process, the prompt has been replaced with the base version.}
    \setlength{\tabcolsep}{4pt}
    \renewcommand{\arraystretch}{1.2}
    \resizebox{\textwidth}{!}{%
    \begin{tabular}{llllcccccccc}
        \toprule
            \textbf{FP-Evol} & \textbf{I-Evol} & \textbf{CR-Evol} & \textbf{I-Elim} &\textbf{\small{MMStar}} & \textbf{\small{MathVista$\rm^{M}$}} & \textbf{\small{MME$\rm^{C}$}} & \textbf{\small{AI2D}} & \textbf{\small{HallBench}} &  \textbf{\small{MMMU$\rm^{V}$}}  & \textbf{\small{RWQA}} & \textbf{\small{AVG.}}\\
            \midrule[0.4pt]
            \cmark & \cmark & \cmark & \cmark & 34.7 & 25.7 & 29.9 & 54.1 & 42.1 & 35.5 & 49.8 & 38.8 \\
            \cmark & \cmark & \fmark & \cmark & 35.7 & 25.9 & 30.3 & 54.8 & 42.9 & 35.2 & 51.2 & 39.4 \\
            \cmark & \fmark & \fmark & \cmark & 36.5 & 25.4 & 30.8 & 55.0 & 43.6 & 35.4 & 52.4 & 39.9 \\
            \fmark & \fmark & \fmark & \cmark & \textbf{37.9} & \textbf{26.1} & \textbf{31.3} & \textbf{55.1} & \textbf{43.8} & \textbf{35.8} & \textbf{53.2} & \textbf{40.5} \\
        \bottomrule
        \end{tabular}
    }
    \vspace{-1pt}
    \label{tab:ablation_prompt}
    \vspace{-2mm}
\end{table}

\subsection{Benchmark Comparison}
After comprehensively validating our approach’s ability to enhance the complexity and diversity of instruction data, we perform a thorough comparison with previous SOTA methods across 13 vision-language benchmarks, summarizing the results in the \cref{tab:sota_comparision}. Notably, we observe that supported by enhanced and refined instruction data, our MLLM significantly advances performance boundaries in almost all benchmarks, consistent with the performance improvements observed in our ablation experiments in \cref{sec:ablation_study}. Remarkably, compared to the fully open-source SOTA model Cambrain-1~\citep{cambrian-10m}, our method, although using seed data sampled from training data of Cambrain-1, achieves superior results with a substantial performance increase ($\uparrow$ 2.9 average points). This indicates that the quality of instruction data is more crucial than quantity.

In comparison to the open-source SOTA model MiniCPM-v2.5~\citep{minicpm}, despite a considerable difference in training data volume, \textbf{MMEvol-8} still delivers better results, particularly showing improvements in instruction following, visual hallucinations, and visual reasoning with gains of $\uparrow$3.1 points on HallBench, $\uparrow$2.5 points on MIA, and $\uparrow$13.6 points on MMSInst respectively. This demonstrates that our method enhances the model's visual reasoning and instruction following, reduces visual hallucinations, and improves other general capabilities, consistent with our findings from ablation studies and qualitative analyses. By using our data and the leading large language model Qwen2, we can train a superior MLLM from scratch in only one day using 4×8 A100 GPUs, further validating that high-quality instruction data is more important than large-scale low-quality data.

\begin{table}[t!]
\centering
\caption{Comparison with state-of-the-art methods on 13 visual-language benchmarks. Our models consistently improve LLaVA-NeXT under a head-to-head comparison, using the same prompts and the same base LLM, showing the effectiveness of enhanced pretraining data quality. ``PT'' denotes pre-training data scale,  ``IT'' denotes instruction tuning data scale and ``*'' denotes the baseline model trained on the seed dataset. We mark the best performance \textbf{bold} and the second-best \uline{underlined}.}
\setlength{\tabcolsep}{4pt}
\renewcommand{\arraystretch}{1.2}
\resizebox{\textwidth}{!}{%
\begin{tabular}{lcccccccccccccccc}
    \toprule
    \textbf{Model} & \rotatebox{90}{\textbf{\small{PT}}} & \rotatebox{90}{\textbf{\small{IT}}}& \rotatebox{90}{\textbf{\small{VQA$\rm^{v2}$}}} & \rotatebox{90}{\textbf{\small{GQA}}} & \rotatebox{90}{\textbf{\small{MME$\rm^C$}}} & \rotatebox{90}{\textbf{\small{MMStar}}} & \rotatebox{90}{\textbf{\small{HallBench}}} & \rotatebox{90}{\textbf{\small{MathVista$\rm^M$}}} & \rotatebox{90}{\textbf{\small{MMMU$\rm^V$}}} & \rotatebox{90}{\textbf{\small{AI2D}}} & \rotatebox{90}{\textbf{\small{POPE}}} & \rotatebox{90}{\textbf{\small{MIA}}} & \rotatebox{90}{\textbf{\small{BLINK}}} & \rotatebox{90}{\textbf{\small{RWQA}}} & \rotatebox{90}{\textbf{\small{MMSInst}}} & \rotatebox{90}{\textbf{\small{AVG.}}} \\     
     \midrule[0.6pt] 
        \multicolumn{17}{l}{\textbf{Weight Open-Source}} \\
        Yi-VL-6B & 125M & 1M & -- & --  & 46.2 & 37.7 & 55.7 & 28.8 & 40.3 & 59.8 & 82.5 & 26.1 & 38.7 & 53.5 & -- & 46.9\\
        DeepSeek-VL-7B & 275M & 50M & -- & --  & 37.1 & 40.5 & 53.9 & 36.8 & 38.3 & 65.3 & 85.6 & 61.0 & 40.9 & 49.7 & 26.7 & 48.7\\
        Qwen-VL-Chat-7B & 1.4B & 50M & 78.2  & 57.5  & 49.0 & 34.5 & 56.4 & 34.9 & 37.0 & 63.0 & 74.9 & 63.1 & 28.2 & 49.3  & -- & 52.2 \\
        CogVLM-Chat-17B & 1.5B & 5.1M & --  & \uline{65.2} & 37.4 & 39.9 & 55.1 & 34.7 & 37.3 & 63.3 & \uline{88.0} & 60.0 & 41.5 & 60.3  & -- & 53.0 \\
        MiniCPM-V2.5-8B & 570M & 9.1M & 81.9 & 64.7 & \uline{50.3} & \uline{51.3} & 59.2 & \textbf{54.3} & \uline{43.0} & \textbf{78.3} & 86.7 & 76.3 & 36.7 & 63.5 & 28.2 & 59.6 \\
        \textcolor{black}{InternVL2-8B}& - & - & - & - & \textcolor{black}{71.8} & \textcolor{black}{61.5} & \textcolor{black}{63.9} & \textcolor{black}{58.3} & \textcolor{black}{51.2} & \textcolor{black}{83.6} & \textcolor{black}{84.2} & 
        - & 
        - & 
        \textcolor{black}{64.2}  & 
        - & 
        \textcolor{black}{67.3}\\
        \textcolor{black}{Qwen2-VL-7B}& - & - & - & - & \textcolor{black}{64.7} & \textcolor{black}{60.7} & \textcolor{black}{68.5} & \textcolor{black}{61.4} & \textcolor{black}{53.7} & \textcolor{black}{83.0} & \textcolor{black}{85.4} & 
        - & 
        - & 
        \textcolor{black}{70.1}  & 
        - & 
        \textcolor{black}{68.4}\\ 
        \midrule[0.6pt] 
        \multicolumn{17}{l}{\textbf{Fully Open-Source}} \\
        InstructBLIP-7B & 0.6M & 0.8M & --  & 49.2  & 31.8 & 32.7 & 53.6 & 24.4 & 30.6 & 40.6 & 86.1 & 38.2 & 39.7 & 36.9 & -- & 42.2 \\
        LLaVA-1.5-7B  & 0.6M & 0.8M & 78.5  & 62.0 & 37.8 & 33.1 & 48.8 & 25.6 & 35.7 & 55.5 & 86.1 & 62.2 & 38.0 & 54.8 &  15.4 & 48.7 \\
        LLaVA-1.5-13B & 0.6M & 0.8M & 80.0  & 63.3 & 34.8 & 34.3 & 45.3 & 27.7 & 37.0 & 61.1 & \textbf{88.4} & 63.6 & 40.9 & 55.3 & -- & 52.6
        \\
        LLaVA-NeXT-8B  & 0.6M & 0.8M & 81.8  & 65.2 & 44.6  & 43.9  & 52.3 & 31.5 & 41.7 & 69.9 & 87.3 & 65.1 & 43.5 & 60.1 & 25.6 & 54.8\\
        LLaVA-NeXT-13B  & 0.6M & 0.8M & 82.8  & 65.4 & 37.1 & 40.4 & 51.5 & 35.1 & 35.9 & 72.2 & 87.8 & 69.2 & 41.2 & 59.1 & 30.2 & 54.5\\
        VILA-1.5-8B & 50.5M & 6.0M & 80.9  & 61.9 & 39.0 & 39.7 & 55.8 & 37.3 & 36.9 & 58.8 & 85.5 & 66.1 & 37.0 & 43.3 & 21.6 & 51.1\\
        VILA-1.5-13B & 50.5M & 6.0M & 82.8  & 64.3  & 38.5 & 44.2 & 59.2 & 42.5 & 37.9 & 69.9 & 84.2 & 61.2 & 41.5 & 53.3  & 30.6 & 54.6\\
        Cambrian-1-8B  & 2.5M & 7.0M & 81.2  & 64.6 & 41.1 & 50.7 & 47.8 & 47.0 & 41.8 & 74.6 & 86.4 & 68.7 & 44.9 & \textbf{64.2} & 28.3 & 57.1\\
        Cambrian-1-13B  & 2.5M & 7.0M & 82.6  & 64.3 & 44.5 & 47.1 & 58.9 & 47.4 & 40.0 & 73.6 & 86.8 & 69.8 & 43.1 & 63.0 & 25.8 & 57.5 \\
        \midrule[0.6pt] 
        \rowcolor{lightgray} \textbf{LLaVA-NeXT*}-8B & 0.6M & 1.1M & 82.5 & 64.8 & 41.3 & 47.4 & 60.8 & 47.7 & 38.0 & 72.1 & 85.3 & 69.4 & 44.2 & 59.9  & 26.2 & 56.9\\ 
        \rowcolor{lightgray} \textbf{LLaVA-NeXT*}-Qwen2-7B & 0.6M & 1.1M & 82.5 & 64.9 & 44.6 & 48.9 & 61.7 & 49.3 & 41.7 & 73.3 & 86.4 & 
        70.2 & 
        44.7 & 
        61.0  & 
        30.1 & 
        58.4 \\ 
        \rowcolor{lightgray} \textbf{MMEvol}-8B & 0.6M & 1.6M & \textbf{83.4} & 65.0  & 47.8 & 50.1 & \uline{62.3} & 50.0 & 40.8 & 73.9 & 86.8 & \textbf{78.8} & \uline{46.4} & 62.6  & \uline{32.3} & \uline{60.0} \\      
        \rowcolor{lightgray} \textbf{MMEvol}-Qwen2-7B & 0.6M & 1.6M & \uline{83.1}  & \textbf{65.5}  & \textbf{55.8} & \textbf{51.6} & \textbf{64.1} & \uline{52.4} & \textbf{45.1} & \uline{74.7} & 87.8 & \uline{77.6} & \textbf{47.7} & \uline{63.9} & \textbf{41.8} & \textbf{62.4}\\   
    \bottomrule
    \end{tabular}
}
\vspace{-10pt}
\label{tab:sota_comparision}
\end{table}

\section{Related Work}

\textbf{Multimodal Large Language Models (MLLMs).} MLLMs have rapidly advanced in recent years due to the success of Large Language Models (LLMs) and the availability of diverse image-text instruction data from the internet. LLaVA~\citep{llava} and MiniGPT-4~\citep{minigpt4} have demonstrated strong cross-task generalization by integrating visual encoders with large language models through simple connectors and training on instruction data. LLaVA-NeXT~\citep{llava-next} has significantly enhanced visual perception by employing dynamic resolution techniques. Cambrain-1~\citep{cambrian-10m} has improved model robustness through visual encoder routing, though it incurs higher training costs. DEEM~\citep{deem} simplifies model architecture and enhances robustness by using diffusion models to extract visual features instead of traditional visual encoders. Subsequent work~\citep{diva,transfusion,showo} following DEEM combine diffusion models with LLMs to further enhance generative and understanding capabilities of MLLMs. However, these models still face challenges related to the quantity and quality of data, which limit performance improvements further.

\textbf{Image-text Instruction Data Construction}. LLaVA~\citep{llava} has improved model capabilities by utilizing LLaVA-Instruct~\citep{llava}, a dataset labeled by advanced LLMs. However, this approach does not fully exploit visual information and have limited instruction types. ALLaVA~\citep{allava}, by manually crafting and rewriting instruction data, offers greater variety but suffers from high manual labeling costs, inefficiency, and overly simplistic problems. MMInstruct~\citep{mminstruct} generates instruction data automatically with advanced MLLMs, but the instruction complexity and diversity are constrained by predefined formats, failing to fully exploit effective visual information. VILA$^2$~\citep{vila2} has generated extensive data through instruction evolution but lacks complexity and variety, limiting its utility for other models. In contrast, we address this challenge and propose \ours, which iteratively enhances instruction diversity and complexity through instruction evolution on limited data, aiming to extract more usable visual information and endow MLLMs with more powerful capabilities.

\section{Conclusions}

In this work, we propose an image-text instruction evolution framework and explore the techniques, insights, and benefits of Evol-Instruct for enhancing the quality and quantity of image-text instruction data. We employ three distinct evolution methods to increase the complexity and diversity of instruction data based on a limited seed dataset while utilizing instruction elimination to filter out harmful data. The data evolved through three rounds of evolution is used to train a new model, demonstrating state-of-the-art (SOTA) performance across a comprehensive set of benchmarks. Future directions include exploring integrating image generation models to synthesize new images and perform dual evolution of images and texts, aiming to train even more robust foundational models.

\section{Limitation}
\vspace{-5pt}
Due to resource limitation, we only performed evolution on 163K samples (approximately 12\% in original data recipes) and conducted experiments with an 8B scale model. Expanding the dataset and using larger-scale models could yield even better results. We plan to explore these avenues in future work and replace the OpenAI GPT4o-mini API with open-sourced model like QWen2VL.


\clearpage

{\small
\bibliographystyle{iclr2025_conference}
\bibliography{iclr2025_conference}
}

\newpage

\appendix
\section*{Appendix}

\section{Curation Details of Seed Data }
\vspace{-5pt}
\label{sec:seed_details}
LLaVA-Instruct~\citep{llava} is a dataset of image-text instructions based on the COCO~\citep{coco} data source and generated using the OpenAI ChatGPT API. The image-text instruction format in this dataset primarily includes three types: dialogue-based question-answering, global descriptions, and complex reasoning. ShareGPT4V~\citep{sharegpt4v}, on the other hand, is a dataset constructed or rewritten using the OpenAI GPT-4V API, based on image-text pairs from SAM~\citep{segany}, COCO, and other sources to introduce richer details into captions. Both LLaVA-Instruct and ShareGPT4V significantly advance the development of MLLMs \citep{unimse,unisa,spokenwoz} and are widely used. We integrate samples from these two datasets containing the same image by concatenating the corresponding instruction data lists. For samples with global descriptions but no instruction data, we use the GPT-4o-mini API to supplement the missing instruction data, similar to LLaVA-Instruct, resulting in a combined dataset of 133K samples. To ensure the diversity of the seed data, we also include additional scientific chart data. Specifically, we sample 30K entries from Cambrain-1~\citep{cambrian-10m}, covering various types of image-text instructions such as code generation, chart interpretation, scientific question-answering, document understanding, and mathematical reasoning, ultimately forming a seed dataset of 163K image-text instructions.

\begin{table*}[h]
    \centering
    \small
    \caption{The mixture of training recipe datasets with corresponding categories and sources. We collect these public dataset form internet.}
    \begin{tabular}{lccc}
        \toprule
        \textbf{Category} & \textbf{Sources} & \textbf{Size} & \textbf{Ratio} \\ 
         \midrule[0.6pt]
       VQA & VQAV2~\citep{vqav2} & 83K & 5.1\%\\
         \midrule[0.6pt]
         \multirow{2}{*}{Knowledge} & OKVQA~\citep{ok-vqa}, A-OKVQA~\citep{aokvqa} & \multirow{2}{*}{243K} & \multirow{2}{*}{14.9\%}\\
         &  VG~\citep{vg}, GeoQA~\citep{geoqa} & & \\
         \midrule[0.6pt]
        Reasoning & GQA~\citep{gqa} & 72K & 4.4\%\\
         \midrule[0.6pt]
         Grounding & RefCOCO~\citep{refercoco} & 48K & 2.9\%\\
         \midrule[0.6pt]
        \multirow{5}{*}{OCR} & OCR-VQA~\citep{ocrvqa}, TextVQA~\citep{textvqa} & \multirow{5}{*}{270K} & \multirow{5}{*}{16.5\%}\\
         &  AI2D~\citep{ai2d}, ChartQA~\citep{chartqa} & & \\
         &  DocVQA~\citep{docvqa}, DVQA~\citep{dvqa} & & \\
         &  Synthdog-EN~\citep{synthdog}, Datikz~\citep{datikz} & & \\
        &  TabMWP~\citep{tabmwp}, ArxivQA~\citep{arxivqa} & & \\
        \midrule[0.6pt]
     Instruct & \ours, ALLaVA~\citep{allava} & 650K & 39.8\%\\
        \midrule[0.6pt]
        Language & ShareGPT, WizardLM~\citep{wizardlm} & 183K & 11.2\%\\
        \midrule[0.6pt]
        \multirow{3}{*}{Science/Code} & Design2Code~\citep{design2code}, MathVision~\citep{mathvision} & \multirow{3}{*}{85K} & \multirow{3}{*}{5.2\%}\\
          &  Geo170k~\citep{geoqa}, ScienceQA~\citep{scienceqa} & & \\
          & Websight~\citep{websight}, Cambrain-Data-Engine~\citep{cambrian-10m} & & \\
        \bottomrule
    \end{tabular}
    \label{tab:recipe}
\vspace{-20pt}
\end{table*}

 \begin{table}[t]
 \vspace{-20pt}
    \centering
    \small
    \caption{The detailed training setup for \ours and the hyper-parameters across the training stages. }
    \begin{tabular}{lcccc}
        \toprule
        \textbf{Hyperparameter} & \textbf{Ablation Stage 1} & \textbf{Ablation Stage 2} & \textbf{SOTA Stage 1} & \textbf{SOTA Stage 2} \\ 
        \midrule[0.4pt]
        \multirow{2}{*}{language model} & LLaMA 3 8b & LLaMA 3 8b & LLaMA 3 8b & LLaMA 3 8b \\
        & & &  Qwen 2 7b & Qwen 2 7b \\
        global batch size & 128 & 128 & 128 & 128 \\
        batch size & 4 & 4 & 4 & 4 \\
        learning rate & 1e-3 & 5e-5 & 1e-3 & 5e-5\\
        lr schedule & cosine & cosine & cosine & cosine \\
        lr warmup ratio & 0.03 & 0.03 & 0.03 & 0.03\\
        weight decay & 0 & 0 & 0 & 0 \\
        epoch & 1 & 1 & 1 & 1 \\
        optimizer & AdamW & AdamW & AdamW & AdamW \\
        cost & 4h & 0.1h & 4h & 20h \\
        dataset & LLaVA Pretrain & Seed-30K/Evol-30k & LLaVA Pretrain & Dataset Mixture \\
        \bottomrule
    \end{tabular}
    \label{tab:hyper-param}
    \vspace{-15pt}
\end{table}

\section{Implementation Details} 
\vspace{-5pt}
\label{sec:recipe_details}
After three rounds of evolution and filtering, we obtain 447K high-quality image-text instruction data with diversity and complexity. This data, combined with the ALLaVA instruction dataset, forms the 600K instruction data segment of the training recipe. To ensure a fair comparison with other methods, we combine the instruction data with other commonly used image-text data into the final training recipe, as shown in the \cref{tab:recipe}. Notably, we find that the DataEngine~\citep{cambrian-10m} data contains many harmful mismatched image-text pairs. We use OpenAI GPT-4o API to filter out harmful data and obtain 20K effective image-text instruction data. More details about training settings can be found in \cref{tab:hyper-param}

\section{Additional Visualization Results}
\label{sec:add_vis}
\begin{figure}[b]
\begin{center}
\includegraphics[width=\linewidth]{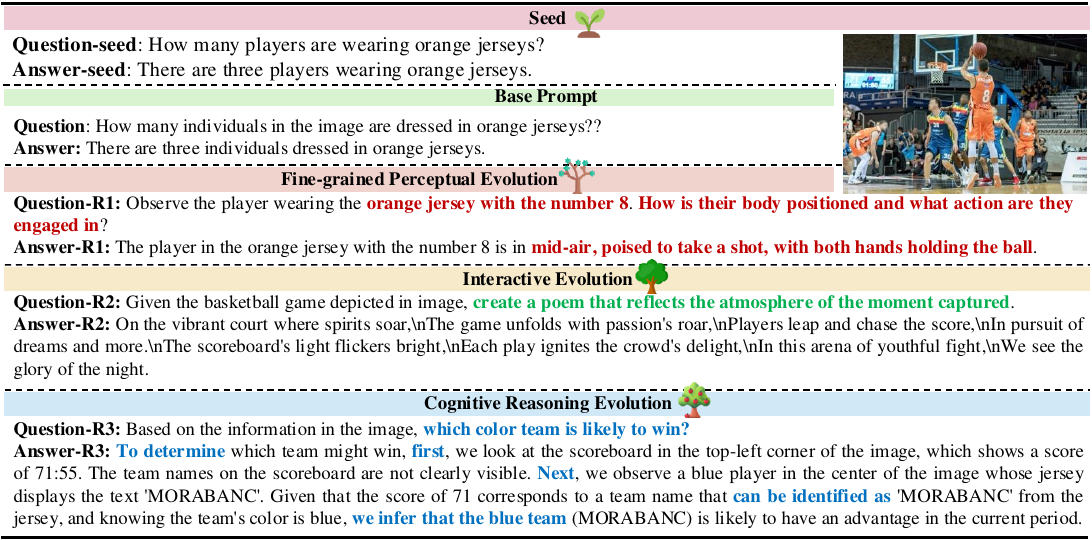}
\end{center}
\caption{\ours continuously enhances instruction data complexity and diversity over evol-instruct. The sample is from SEED-163K. We mark fine-grained visual information in red, new instructions form in green, and longer reasoning steps in blue. \textcolor{black}{Additionally, we also provide the rewritten samples via a base prompt for comparison.}}
\label{fig:case}
\vspace{-0.4cm}
\end{figure}

We sample a specimen from SEED-163K and display its evolution process in \cref{fig:case}. In round 1, we perform fine-grained perceptual evolution, leading to instruction data with more precise details, including actions and attributes. In round 2, interaction evolution shifts instruction forms from general question answering to creative poetry generation, increasing the diversity of instruction formats. In round 3, cognitive reasoning evolution adds reasoning steps to the answers in the instruction data, enhancing its complexity. Through multiple rounds of instruction evolution, we improve the diversity and complexity of the seed data.

\begin{figure}[t]
\begin{center}
\includegraphics[width=\linewidth]{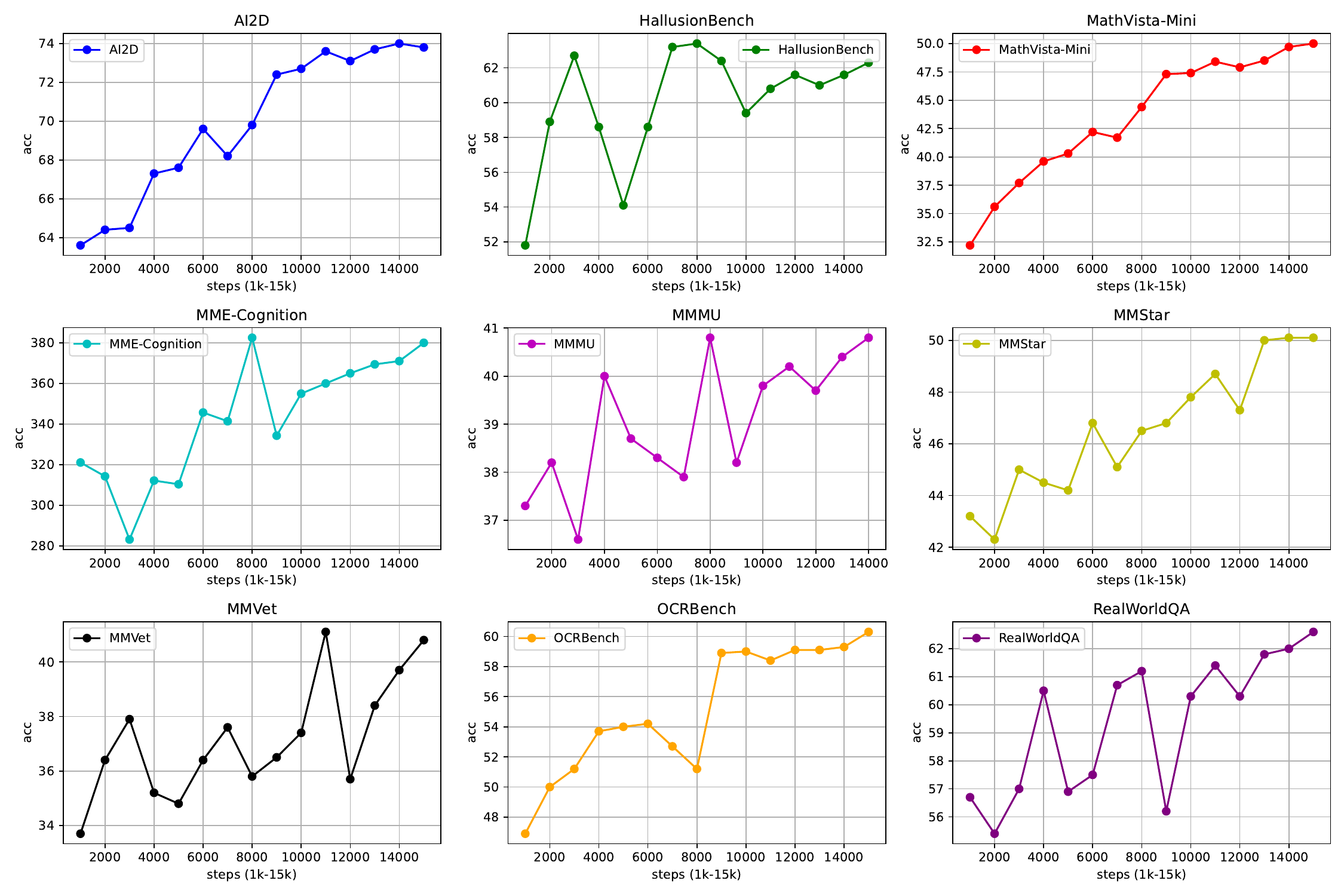}
\end{center}
\caption{\textbf{Illustration of Performance Trends Across Various Benchmarks.} We plot the performance of the model at every 1k step across 9 evaluation datasets to observe the learning trends during training.}
\label{fig:ckpt_performance}
\end{figure}

\begin{figure}[t]
\begin{center}
\includegraphics[width=\linewidth]{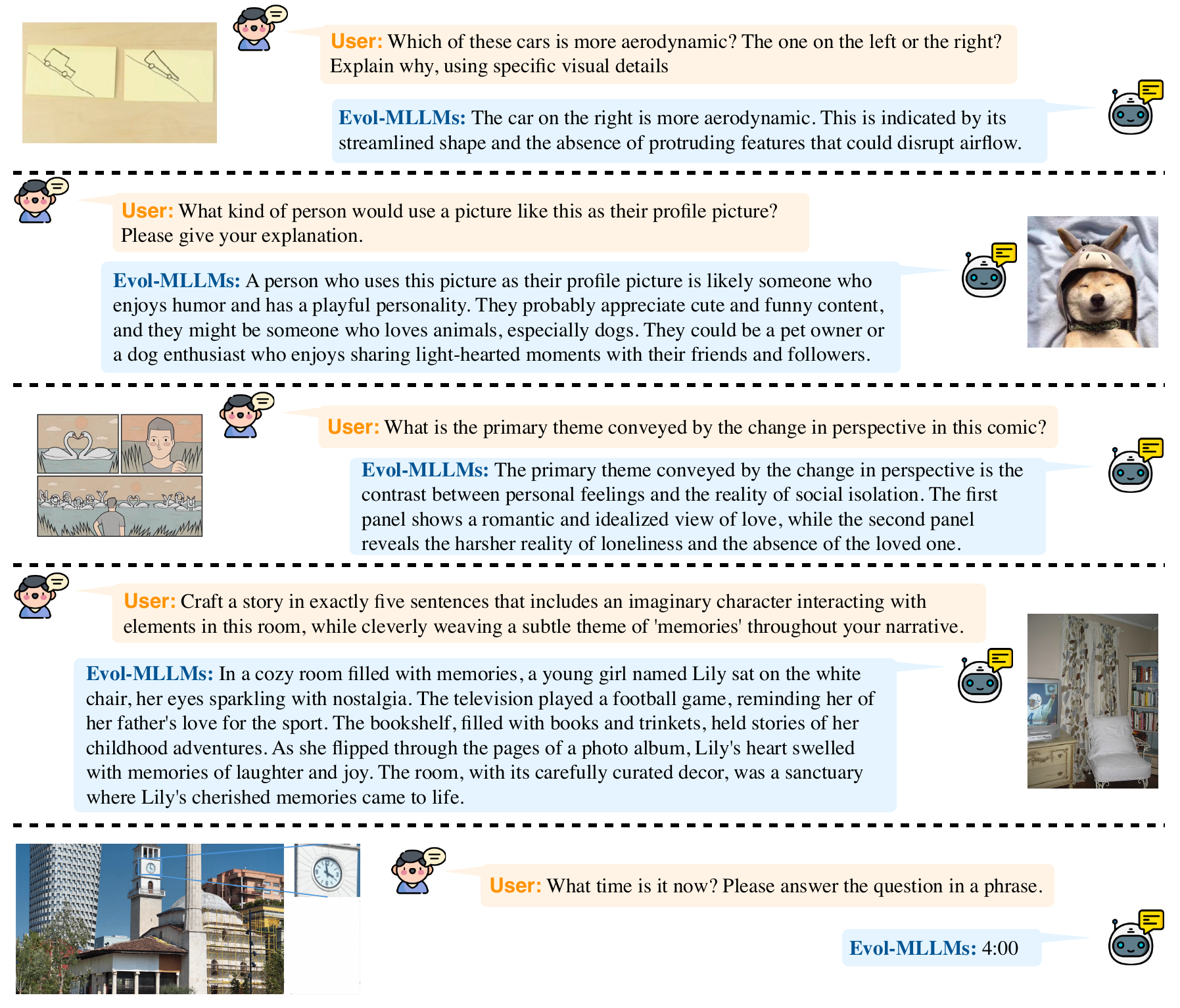}
\end{center}
\caption{\textbf{Examples of image-text dialogue with our MMEvol-8B Model.} Training our model on the enhanced data not only enables it to handle handwritten mathematical reasoning, complex instruction following, and the understanding of emotional nuances, but also allows it to perceive fine-grained information within images, demonstrating the effectiveness of our approach.}
\label{fig:supp_vis}
\end{figure}

\begin{table}[ht]
    \centering
    \caption{\textbf{Human Evaluation of Instruction Evolution.} The table shows the accuracy of manual evaluations by five experts on \ours's evolution and filtering in three directions.}
    \setlength{\tabcolsep}{4pt}
    \renewcommand{\arraystretch}{1.2}
    \resizebox{\textwidth}{!}{%
    \begin{tabular}{cccccccc}
        \toprule
        \textbf{Data ID} & \textbf{Expert} & \textbf{Image Categories} & \textbf{FP-Evol (0-5)} & \textbf{I-Evol (0-5)} & \textbf{CR-Evol (0-5)} & \textbf{I-Elim (0-15)} \\
        \midrule
        0,1,3,4,5,6 & 0 & LandMark, OCR, Human \& Clothes, Traffic, Living room, Sport & 5,4,4,5,5,4 & 5,4,3,4,5,4 & 5,3,4,5,4,4 & 15,13,13,14,13,14 \\
        7,8,9,10,11,12 & 1 & Kitchen, Office supplies \& Tools, Plants, Animal, Sport, LandMark & 5,5,4,5,4,4 & 5,4,5,5,4,4 & 5,5,4,4,5,4 & 14,15,13,15,14,13 \\
        13,14,15,16,17,18 & 2 & Foods, LandMark, OCR, Human \& Clothes, Traffic, Sport & 4,4,3,5,4,5 & 5,4,4,4,4,5 & 4,5,5,4,5,5 & 14,14,15,13,14,15 \\
        19,20,21,22,23,24 & 3 & Foods, Sport, LandMark, Office supplies \& Tools, Plants, Traffic & 3,4,5,5,5,4 & 3,4,5,5,5,5 & 5,5,5,5,5,5 & 13,15,14,15,15,15 \\
        25,26,27,28,29,30 & 4 & Animal, Sport, Traffic, Landmark, Sport, Office supplies \& Tools & 4,5,5,5,5,5 & 4,5,5,5,4,5 & 5,5,3,5,5,5 & 14,15,14,15,14,15 \\
        \midrule
        \multicolumn{3}{c}{\textbf{Average Scores}} & 89.3\% & 88.7\% & 92\% & 94.5\% \\
        \bottomrule
    \end{tabular}
    }
    \vspace{-1pt}
    \label{tab:human}
    \vspace{-2mm}
\end{table}
We plot the performance of the model at every 1k step across 9 evaluation datasets in \cref{fig:ckpt_performance} to observe the learning trends during training. We can observe that the model learns OCR-related capabilities and mathematical reasoning abilities relatively smoothly, while general perception and cognitive skills exhibit more challenges. This may stem from conflicts arising from multi-source training tasks. A phased learning approach based on the difficulty of different tasks could be adopted to achieve better performance. We also present additional visualization results to demonstrate the capabilities of our model. As shown in \cref{fig:supp_vis}, our model trained on this data exhibits strong visual reasoning, instruction following, and fine-grained perception capabilities. Additionally, it identifies nuances in meme content, validating the effectiveness and efficiency of \ours.

\begin{figure}[t]
\begin{center}
\includegraphics[width=\linewidth]{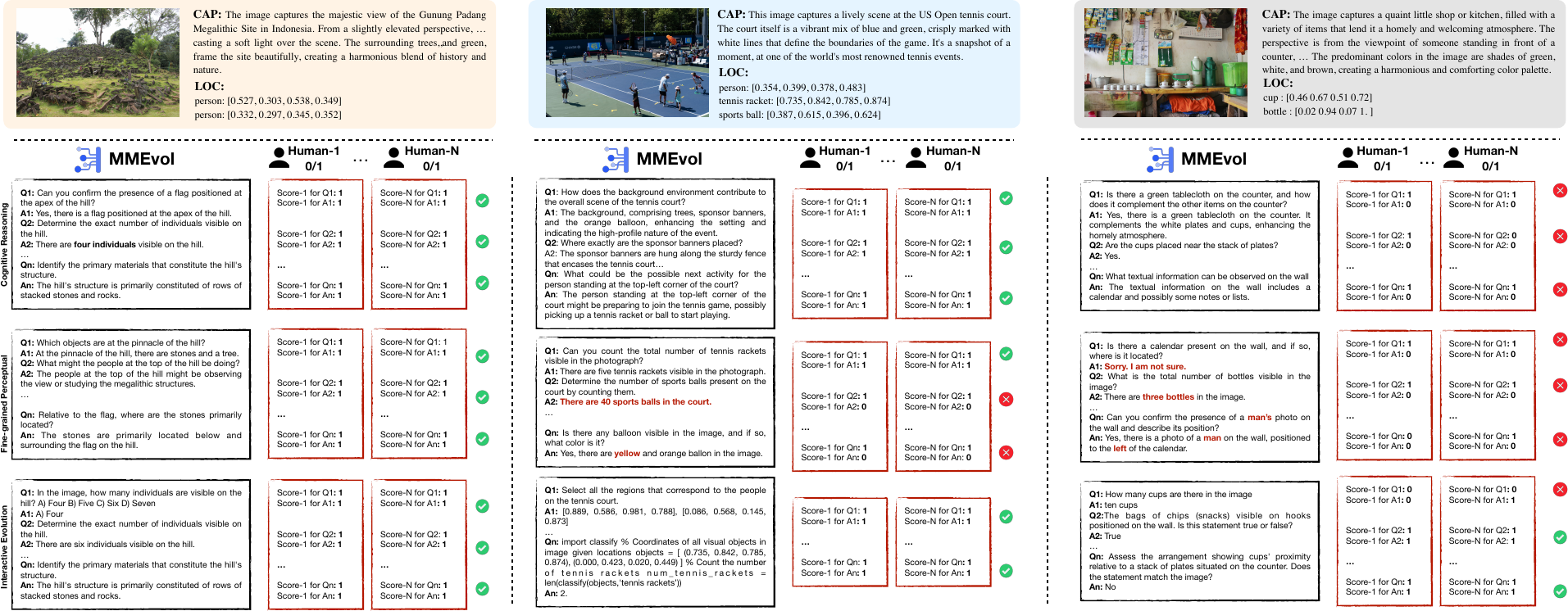}
\end{center}
\caption{\textcolor{black}{\textbf{Schematic diagram of the manual filtering process.} We hired N=5 experts to score (0/1) each question and answer. In the event that any question or answer receives a score of 0, the entire QA pair will be deemed invalid and discarded.}}
\label{fig:human_anno}
\end{figure}

To investigate the reliability of the rewrites produced by GPT-4-o-mini, we conducted a manual evaluation of the data before and after the evolution process. Specifically, we first extracted 30 images of various types from the seed data to ensure diversity, keeping 5 relevant question-answer pairs for each image. Subsequently, we carried out the corresponding evolution in three different directions, ultimately obtaining 450 evolved question-answer pairs, which were then subject to scoring and filtering. The results were distributed among five experts for manual evaluation of the accuracy of the model evolution and the scoring filter. The data is summarized in the \cref{tab:human}. From the table, it is evident that the average success rate of evolution using MLLM can reach 90\%, while the accuracy of the scoring filter can achieve 94\%, indicating the reliability of MMEovel. Additionally, we provide detailed scoring cases in \cref{fig:human_anno}, highlighted in red.
\vspace{-17pt}

\begin{table}[ht]
    \centering
    \caption{\textbf{Ablation study on VLM for instruction evolution.} If we utilize a more advanced open-source model such as Qwen2VL with 72 billion parameters, \ours could achieve further improvements.}
    \setlength{\tabcolsep}{4pt}
    \renewcommand{\arraystretch}{1.2}
    \resizebox{\textwidth}{!}{%
    \begin{tabular}{lcccccccc}
        \toprule
            \textbf{VLM} &\textbf{\small{MMStar}} & \textbf{\small{MathVista$\rm^{M}$}} & \textbf{\small{MME$\rm^{C}$}} & \textbf{\small{AI2D}} & \textbf{\small{HallBench}} &  \textbf{\small{MMMU$\rm^{V}$}}  & \textbf{\small{RWQA}} & \textbf{\small{AVG.}}\\
            \midrule[0.4pt]
            GPT4o-mini (3K) & 37.9 & 26.1 & 31.3 & 55.1 & 43.8 & 35.8 & \textbf{53.2} & 40.5 \\
            Qwen2VL-72B (3K) & \textbf{39.1} & \textbf{27.9} & \textbf{33.1} & \textbf{57.8} & \textbf{46.4} & \textbf{36.9} & 46.9 & \textbf{41.2} \\
        \bottomrule
        \end{tabular}
    }
    \vspace{-1pt}
    \label{tab:ablation_api}
    \vspace{-2mm}
\end{table}

\section{Additional Ablation Study}
\label{sec:add_ablation}

As shown in the \cref{tab:ablation_api}, we achieve better results by replacing the closed-source model with the more advanced open-source Qwen2-VL, demonstrating the scalability and promise of our method. Furthermore, to fully showcase our technological contribution, we use the state-of-the-art method MIMIC-IT~\cite{li2023mimic} to construct a dataset under the same seed data and API conditions, and conduct experiments. The results, as shown in the \cref{tab:ablation_method}, indicate that our method significantly outperforms MIMIC-IT. Additionally, our method is not restricted by data format or limited task forms, making it more user-friendly. We also perform robustness testing on the initial data of our method by selecting samples with a complexity score below 5 to construct a low-quality 1k seed dataset and conduct three rounds of evolution. As shown in the \cref{tab:ablation_init}, \ours is affected by low-quality initial seed samples, resulting in slightly lower outcomes compared to randomly sampled initial seed data. However, after multiple rounds of evolution, the gap narrows significantly, indicating \ours's strong robustness and ability to quickly adapt and generate more complex and diverse high-quality samples.

\section{Complete Evolution Prompt Template}
\label{sec:add_prompt_details}
\begin{figure}[t]
\begin{center}
\includegraphics[width=\linewidth]{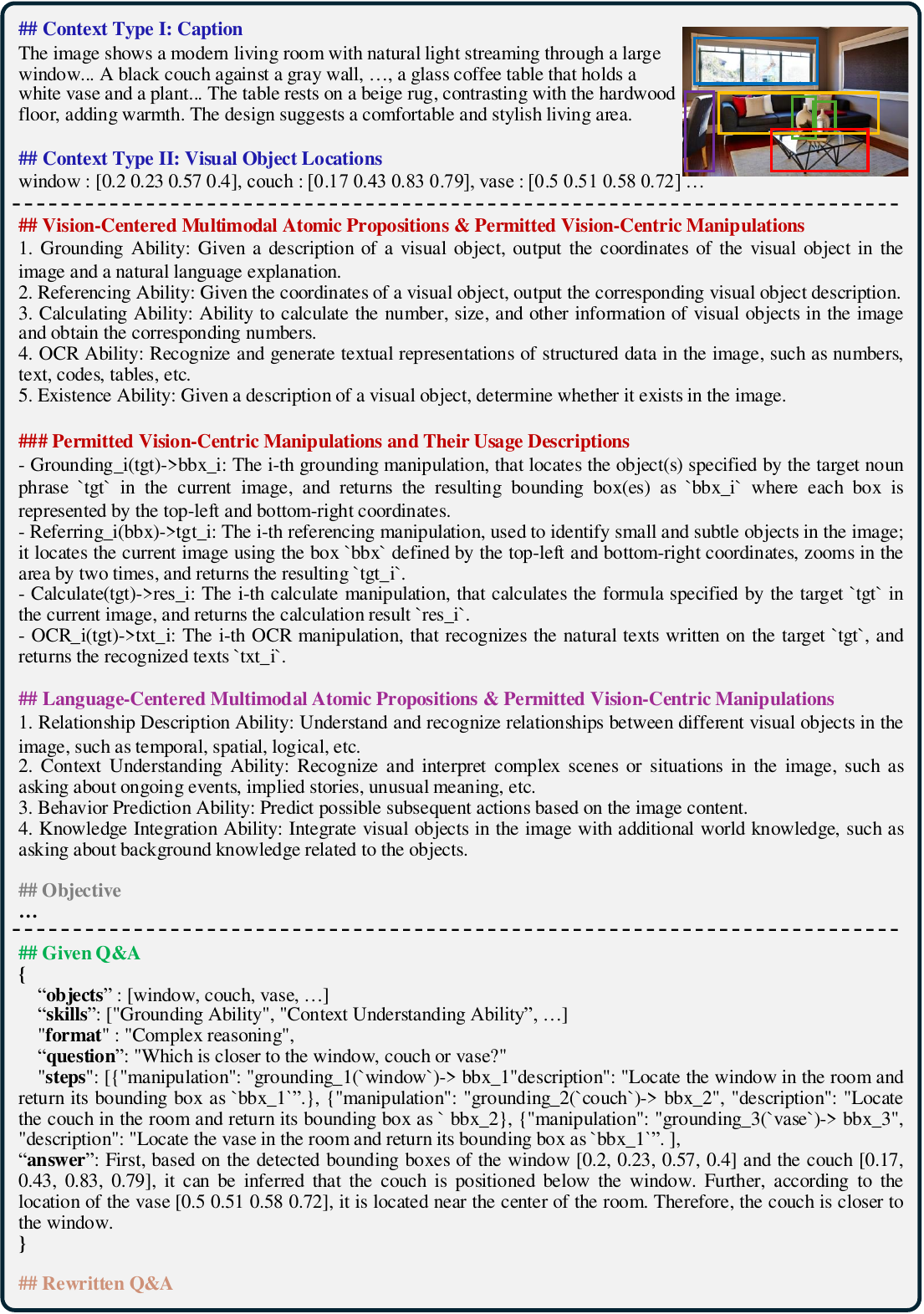}
\end{center}
\caption{\textbf{Complete prefix-prompt template of \ours.}}
\label{fig:complete_head}
\end{figure}

\begin{figure}[t]
\begin{center}
\includegraphics[width=\linewidth]{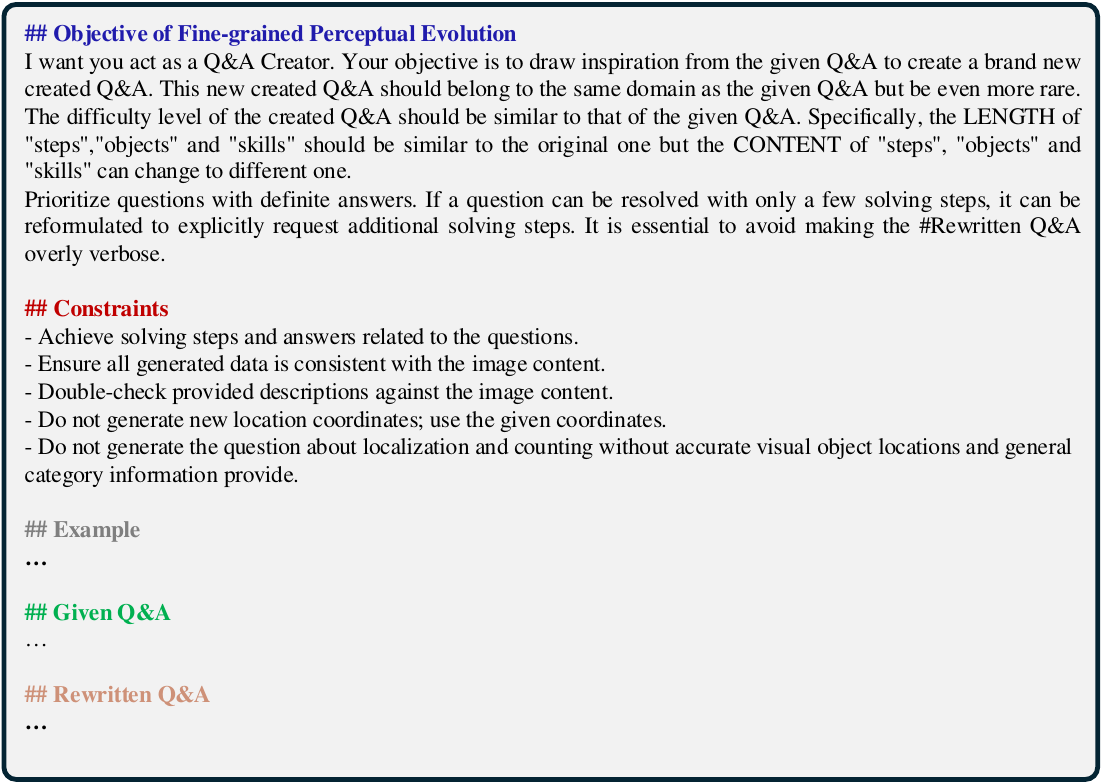}
\end{center}
\caption{\textbf{Complete fine-grained perceptual evolution prompt template.}}
\label{fig:complete_FPE}
\end{figure}

\begin{figure}[t]
\begin{center}
\includegraphics[width=\linewidth]{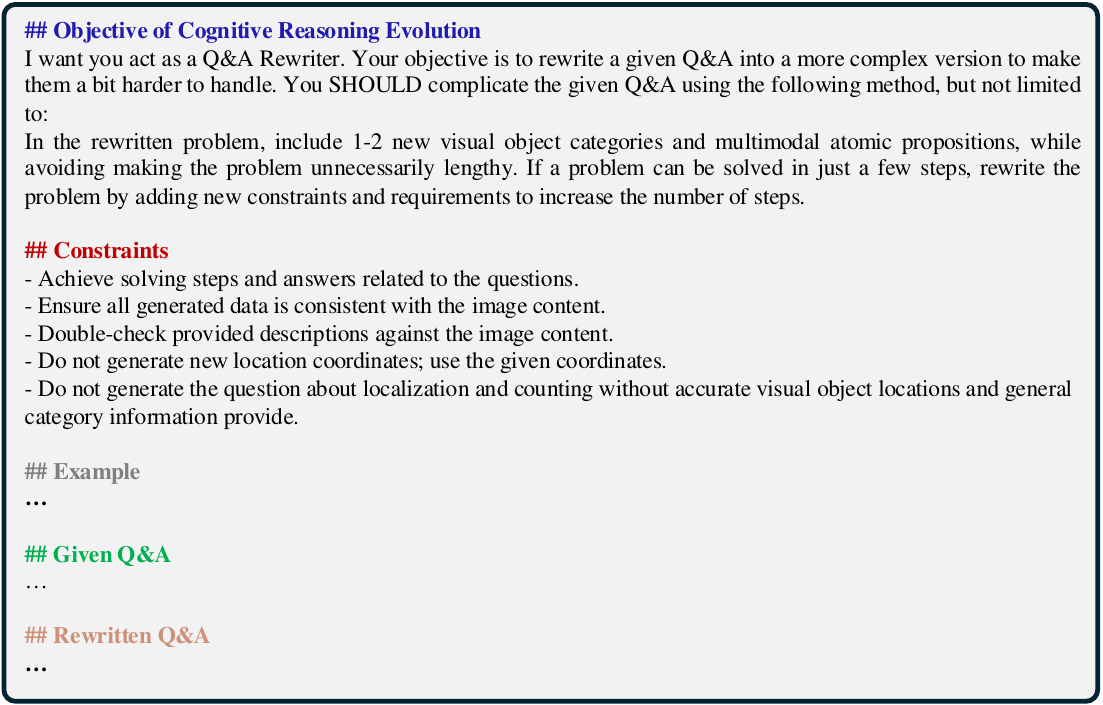}
\end{center}
\caption{\textbf{Complete cognitive reasoning evolution prompt template.}}
\label{fig:complete_CRE}
\end{figure}

\begin{figure}[t]
\begin{center}
\includegraphics[width=0.97 \linewidth]{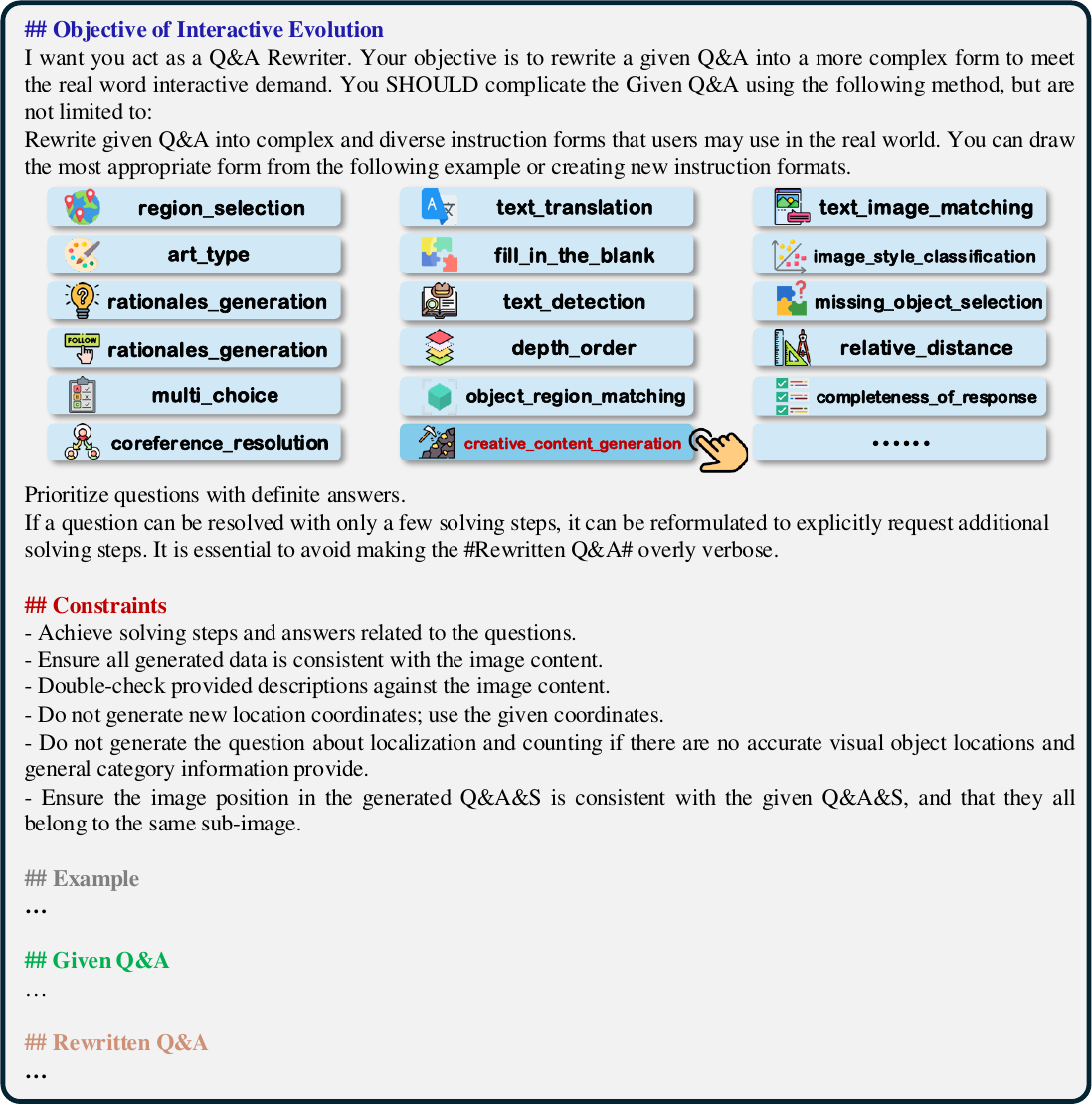}
\end{center}
\vspace{-2mm}
\caption{\textbf{Complete interactive evolution prompt template.}}
\label{fig:complete_IE}
\end{figure}

\begin{figure}[t]
\begin{center}
\includegraphics[width=0.97 \linewidth]{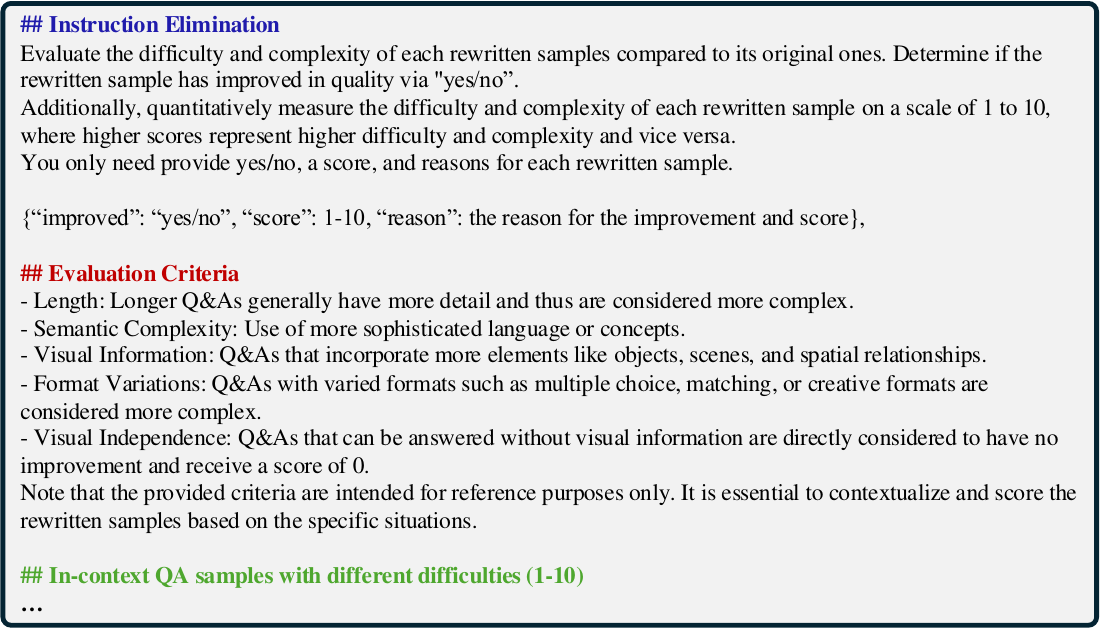}
\end{center}
\vspace{-2mm}
\caption{\textbf{Complete instruction elimination prompt template.}}
\vspace{-5mm}
\label{fig:complete_IE_filter}
\end{figure}

Due to the space limitations in the main text, we simplify the instruction evolution prompt template. We provide the complete detailed evolution templates as follows: the complete prefix-prompt template is shown in \cref{fig:complete_head}, the fine-grained perception evolution prompt template is in \cref{fig:complete_FPE}, the cognitive reasoning evolution prompt template is in \cref{fig:complete_CRE}, the interaction evolution prompt template is in \cref{fig:complete_IE}, and the instruction elimination prompt template is in \cref{fig:complete_IE_filter}.

\begin{table}[ht]
    \centering
    \caption{\textbf{Ablation study on Method.} \ours demonstrates superior performance compared to the previous state-of-the-art method, MIMIC-IT, under the same conditions, validating the effectiveness and efficiency of iterative instructions with explicit optimization objectives.}
    \setlength{\tabcolsep}{4pt}
    \renewcommand{\arraystretch}{1.2}
    \resizebox{\textwidth}{!}{%
    \begin{tabular}{lcccccccc}
        \toprule
            \textbf{Method} &\textbf{\small{MMStar}} & \textbf{\small{MathVista$\rm^{M}$}} & \textbf{\small{MME$\rm^{C}$}} & \textbf{\small{AI2D}} & \textbf{\small{HallBench}} &  \textbf{\small{MMMU$\rm^{V}$}}  & \textbf{\small{RWQA}} & \textbf{\small{AVG.}}\\
            \midrule[0.4pt]
            MIMIC-IT (3K) & 32.1 & 24.3 & 26.4 & 47.6 & 41.9 & 31.5 & 34.5 & 34.1 \\
            \ours (3K) & \textbf{37.9} & \textbf{26.1} & \textbf{31.3} & \textbf{55.1} & \textbf{43.8} & \textbf{35.8} & \textbf{53.2} & \textbf{40.5} \\
        \bottomrule
        \end{tabular}
    }
    \vspace{-1pt}
    \label{tab:ablation_method}
    \vspace{-2mm}
\end{table}

\begin{table}[ht]
    \centering
    \caption{\textbf{Ablation study on Initialization.} The quality of evolutionary data is influenced by the initial instructions, although the impact is relatively minor. Nevertheless, high-quality instructional data can still be generated through multiple iterations, demonstrating the robustness of our method.}
    \setlength{\tabcolsep}{4pt}
    \renewcommand{\arraystretch}{1.2}
    \resizebox{\textwidth}{!}{%
    \begin{tabular}{lcccccccc}
        \toprule
            \textbf{Init} &\textbf{\small{MMStar}} & \textbf{\small{MathVista$\rm^{M}$}} & \textbf{\small{MME$\rm^{C}$}} & \textbf{\small{AI2D}} & \textbf{\small{HallBench}} &  \textbf{\small{MMMU$\rm^{V}$}}  & \textbf{\small{RWQA}} & \textbf{\small{AVG.}}\\
            \midrule[0.4pt]
            Low Score & 36.5 & 25.4 & 29.9 & 54.8 & 43.1 & 35.0 & 52.6 & 39.7 \\
            Random & \textbf{37.9} & \textbf{26.1} & \textbf{31.3} & \textbf{55.1} & \textbf{43.8} & \textbf{35.8} & \textbf{53.2} & \textbf{40.5} \\
        \bottomrule
        \end{tabular}
    }
    \vspace{-1pt}
    \label{tab:ablation_init}
    \vspace{-2mm}
\end{table}

\begin{table}[ht]
    \vspace{-5pt}
    \centering
    \small
    \setlength{\tabcolsep}{1pt}
    \caption{Benchmarks for evaluation with their sources and tested skills. The names are abbreviated due to space limitations. VQA$\rm ^{V2}$; GQA; VQA$\rm^T$: TextVQA; MME$\rm^C$: MME-Cognition; MathVista$\rm^M$: MathVista-MINI; MMMU; AI2D; POPE; HallusionBench: HallBench; MIA; BLINK; RWQA: RealWorldQA; MMSInst: MM-Self-Instruct.}
    \begin{tabular}{cccc}
        \toprule
        \textbf{Skills} & \textbf{Sources} & \textbf{Skills} & \textbf{Sources} \\
        \midrule[0.4pt]
        VQA & VQA$\rm^{v2}$~\citep{vqav2} & \multirow{2}{*}{General Knowledge} & MME$\rm^C$~\citep{mme} \\ 
        \cmidrule(r){1-2}
        Knowledge Leakage & MMStar~\citep{mmstar} & & MMMU~\citep{mmmu} \\
        \midrule[0.4pt]
        Math Reasoning &  MathVista$\rm^M$~\citep{mathvista} &       \multirow{2}{*}{Hallucination} &  POPE~\citep{pope}\\
        \cmidrule(r){1-2}
        OCR Related &  AI2D~\citep{ai2d} & & HallBench ~\citep{hallusionbench}\\
        \midrule[0.4pt]
        Instruction Following & MIA~\citep{miabench} & \multirow{2}{*}{Visual Reasoning} & GQA~\citep{gqa} \\ 
        \cmidrule(r){1-2}
        Visual Perception & BLINK~\citep{blink}, RWQA & & MMSInst ~\citep{mm-cot} \\
        \bottomrule
    \end{tabular}
    \label{tab:benchmark}
    \vspace{-0.4cm}
\end{table}

\begin{figure}[t]
\begin{center}
\includegraphics[width=\linewidth]{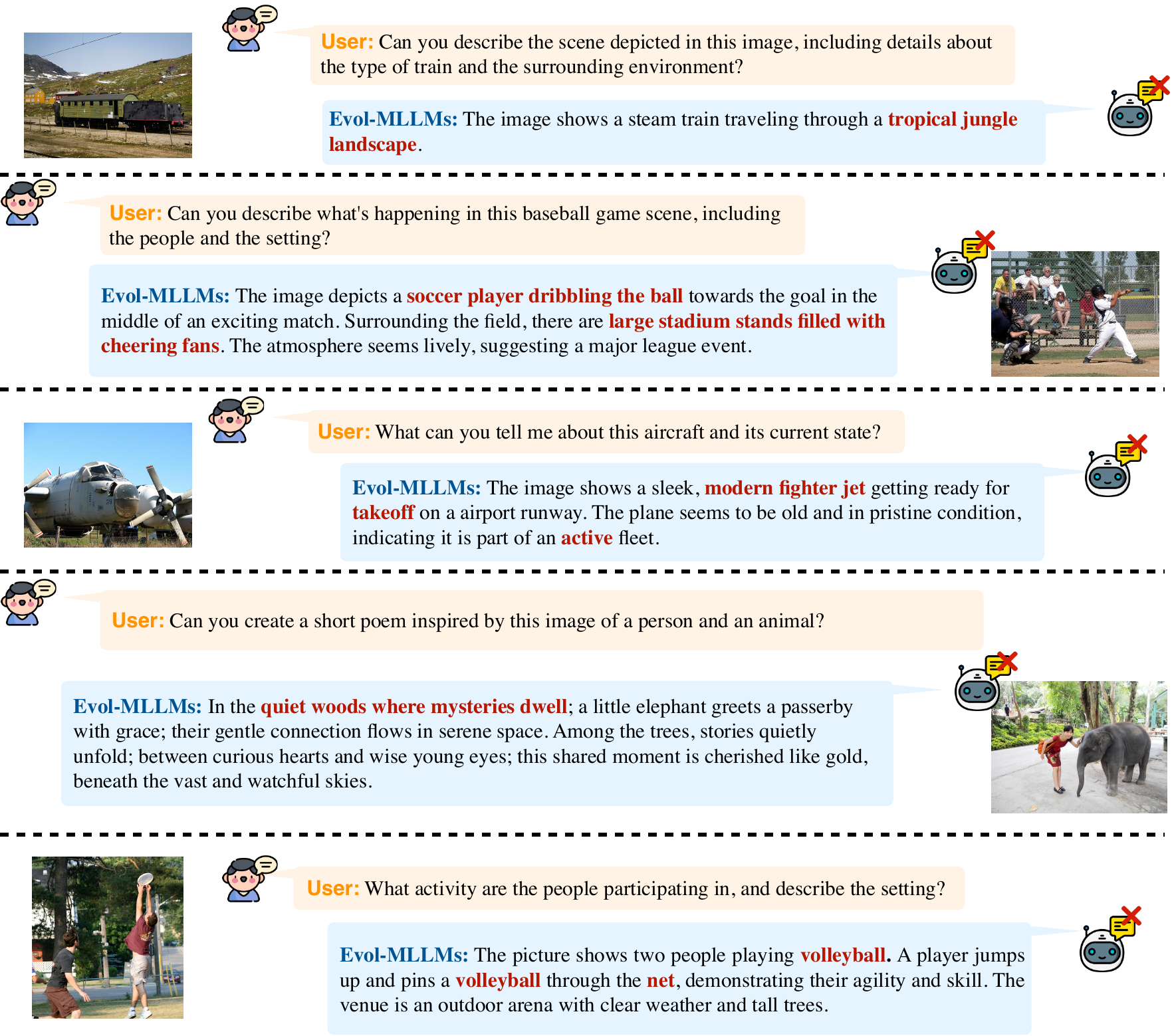}
\end{center}
\caption{\textcolor{black}{\textbf{Some failed cases of image-text dialogue with our MMEvol-8B Model.} Our model generalizes to multiple visual reasoning scenes when
training on the instruction data via our evolution strategy. However, there can
still be errors in some scenarios, \eg,  hallucination.}}
\label{fig:bad_case}
\end{figure}

\begin{figure}[t]
\begin{center}
\includegraphics[width=\linewidth]{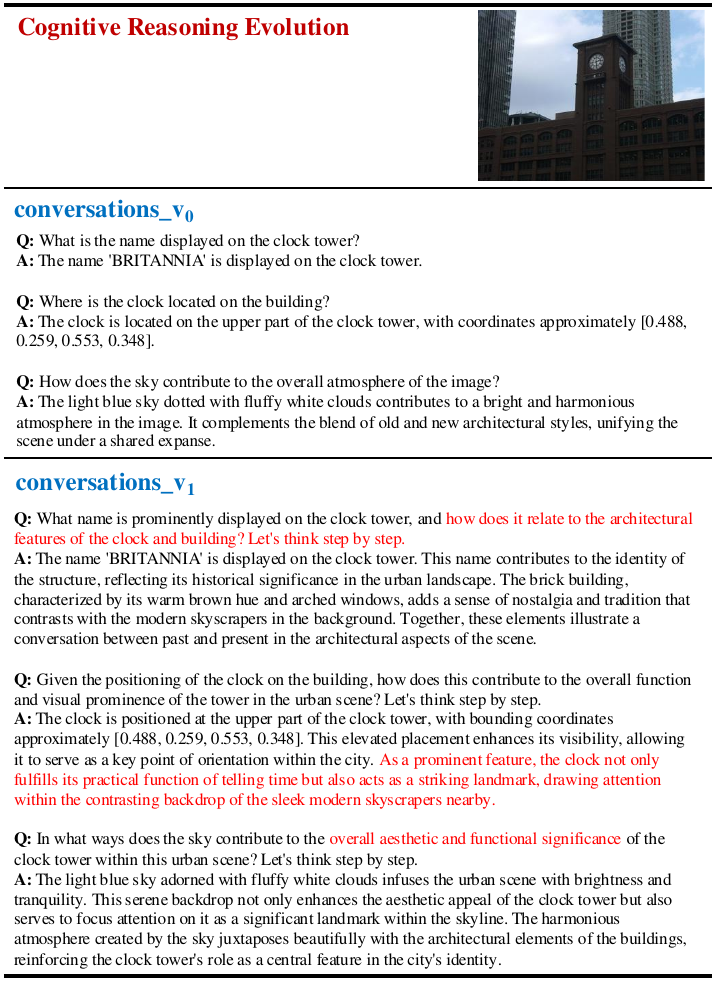}
\end{center}
\caption{\textcolor{black}{\textbf{Data case of cognitive reasoning evolution.}}}
\label{fig:cre_case}
\end{figure}

\begin{figure}[t]
\begin{center}
\includegraphics[width=\linewidth]{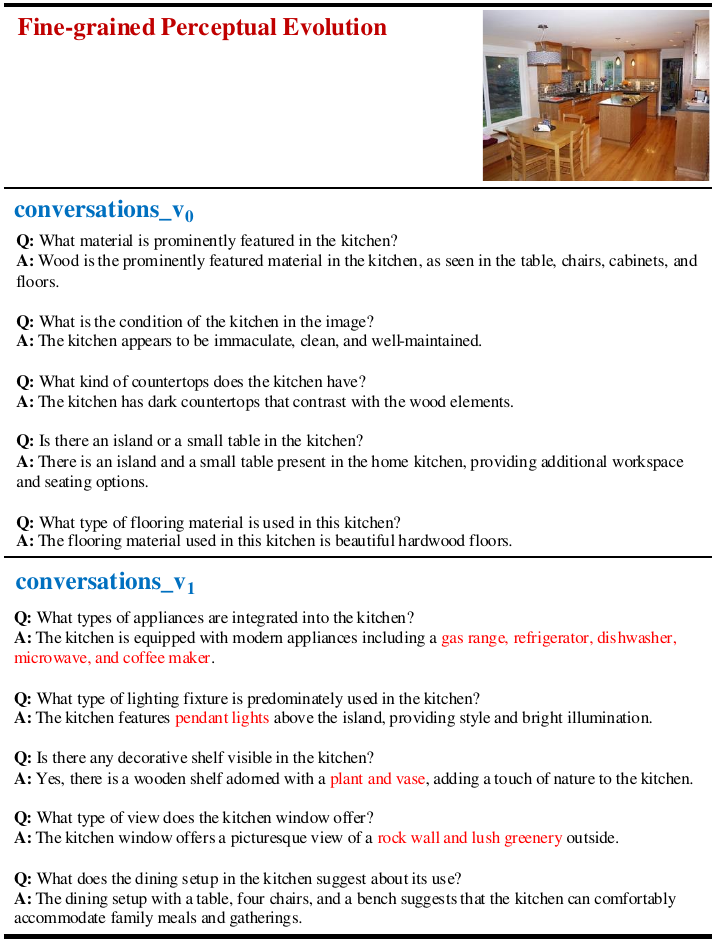}
\end{center}
\caption{\textcolor{black}{\textbf{Data case of fine-grained perceptual evolution.}}}
\label{fig:fpe_case}
\end{figure}

\begin{figure}[t]
\begin{center}
\includegraphics[width=\linewidth]{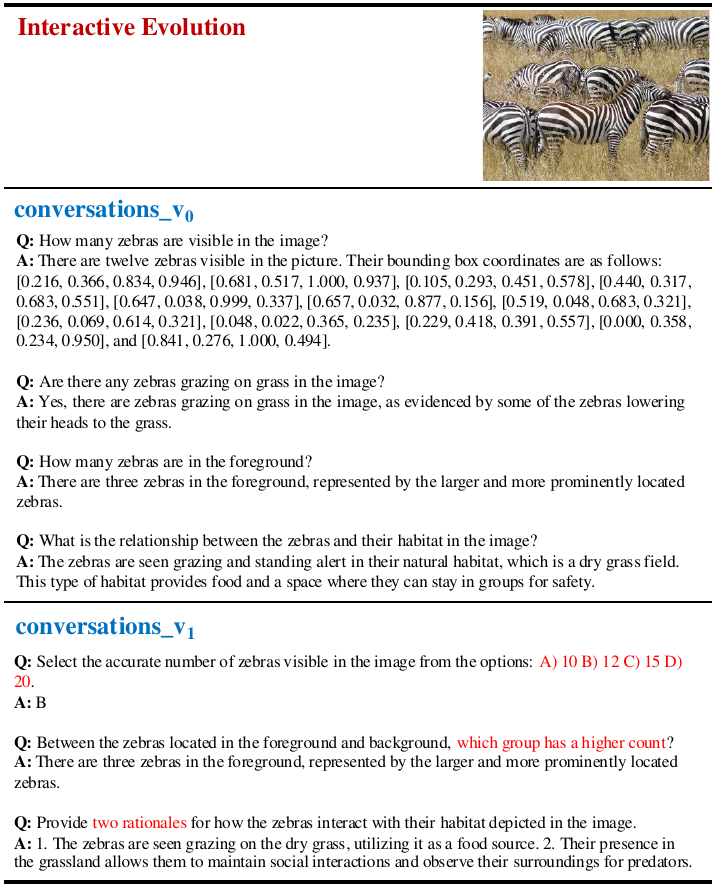}
\end{center}
\caption{\textcolor{black}{\textbf{Data case of interactive evolution.}}}
\label{fig:ie_case}
\end{figure}

\end{document}